\documentclass[10pt,twocolumn,letterpaper]{article}

\usepackage[pagenumbers]{cvpr} 
\usepackage{graphicx}
\usepackage{booktabs}
\usepackage{colortbl}
\usepackage{makecell}
\usepackage{tikz,pgfplots,pgfplotstable}
\usepackage{color}
\usepackage{amsfonts}
\usepackage{amssymb}
\usepackage{arydshln}
\usepackage{multirow}
\usepackage{tcolorbox}

\definecolor{cvprblue}{rgb}{0.21,0.49,0.74}
\definecolor{object_skyblue}{rgb}{0.405, 0.794, 0.902}
\definecolor{aquamarine}{rgb}{0.4, 1.0, 0.6}
\definecolor{ocr_coral}{rgb}{0.749,0.565,0}
\definecolor{hoi_green}{rgb}{0.492,0.671,0.333}
\usepackage[pagebackref,breaklinks,colorlinks,allcolors=cvprblue]{hyperref}

\def\paperID{***} 
\def\confName{CVPR}
\def\confYear{2026}

\title{Enhancing Descriptive Captions with Visual Attributes for Multimodal Perception}

\author{Yanpeng Sun$^{1,2,5}$, \qquad Jing Hao$^{3}$ , \qquad Ke Zhu$^{4}$, \qquad Jiang-Jiang Liu$^{2}$, \qquad Yuxiang Zhao$^{2}$ \\ \qquad Xiaofan Li$^{2}$, \qquad Na Zhao$^{5}$, \qquad Zechao Li$^{1\dagger}$, \qquad Jingdong Wang$^{2}$\thanks{Corresponding author.}\\[3mm]
$^1$NJUST,\qquad $^2$Baidu VIS,\qquad $^3$HKU ,\qquad $^4$NJU, \qquad $^5$SUTD \\
{\tt\small \{yanpeng\_sun, zechao.li\}@njust.edu.cn}
}

\begin{document}
\maketitle
\begin{abstract}
Training Large Multimodality Models (LMMs) relies on descriptive image caption that connects image and language. Existing methods for generating such captions often rely on distilling the captions from pretrained LMMs, constructing them from publicly available internet images, or even generating them through human annotation. However, these strategies can fall short in terms of precision and granularity, particularly when dealing with complex visual reasoning tasks. In this paper, we propose to leverage off-the-shelf visual specialists, which were trained from annotated images initially not for image captioning, for enhancing the image caption. Our approach, named EDC, explores object low-level and fine-grained attributes (e.g., depth, emotion and fine-grained categories) and object relations (e.g., relative location and human-object-interaction (HOI)), and combine the attributes into the descriptive caption. By systematically integrating these rich attributes into the generated captions, EDC significantly improves the descriptive quality of the captions, providing a deeper and more nuanced understanding of the visual content. Experiments demonstrate that such visual specialists are able to improve the performance for visual understanding tasks as well as reasoning that benefits from more accurate visual understanding. 
\end{abstract}    
\section{Introduction}
Recent advancements in Large multimodal models (LMMs)~\cite{wang2023cogvlm,zhu2024self} have significantly enhanced the understanding and reasoning abilities for multimodal tasks. 
Vision-language connection is crucial for high abilities, and descriptive image captions serve as one of key components for image perception~\cite{yu2019multimodal,ji2022knowing}. The image caption is expected to describe the image as detailed and complete as possible. There are two main categories for image captions. One is to generate image captions from human annotation~\cite{feng2012automatic,zha2019context}, such as COCO~\cite{lin2014coco}and LAION~\cite{laion5b}. However, the high cost of human annotation limits scalability. The other one is LMMs~\cite{achiam2023gpt4} annotation, such as ShareGPT4V~\cite{chen2023sharegpt4v} and Densefusion~\cite{Densefusion}. While captions generated by LMMs offer better scalability, their comprehensiveness and accuracy often fall short.

\begin{figure*}
    \centering
    \vspace{-.5em}
    \includegraphics[width=\linewidth]{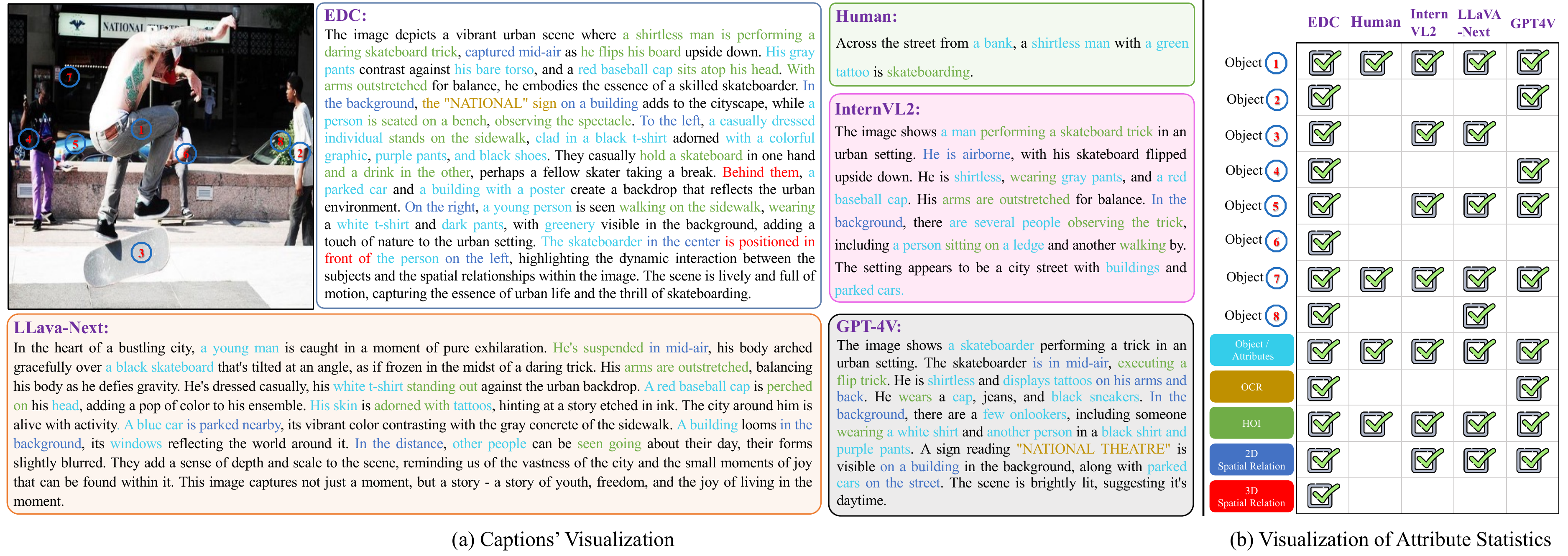}
    \vspace{-1.8em}
    \caption{(a) We present a comparison of captions from EDC, human, and generalist LMM models annotations, including InternVL2-26B, LLaVA-NeXT, and GPT-4V. (b) visualizes the extent to which the captions in (a) describe multiple objects and various attributes, including Objects 1-8, \textcolor{object_skyblue}{Object Attributes}, \textcolor{ocr_coral}{OCR}, \textcolor{hoi_green}{HOI}, \textcolor{blue}{2D spatial relations} and \textcolor{red}{3D spatial relations}. 
    }
    \label{fig:drawbacksofcurrentcaptions}
    \vspace{-2.em}
\end{figure*}

There is still much improvement space for the captions from SoTA LMMs, such as InternVL2~\cite{chen2024internvl} and from current human annotation datasets, which we illustrate through concrete examples in Figure~\ref{fig:drawbacksofcurrentcaptions}. First, human-annotated captions tend to be incomplete, often describing only the most salient objects while omitting finer-grained details and contextual relationships. For example, in Figure~\ref{fig:drawbacksofcurrentcaptions}(a), the widely used COCO dataset provides human annotations that capture only a part of the scene. These captions typically focus on high-level descriptions (\textit{e.g.}, “a shirtless man with a green tattoo is skateboarding”) while ignoring crucial elements such as the number of objects, their fine-grained attributes, or spatial relationships. This oversimplification limits the utility of such captions, particularly in enhancing the vision-language alignment of LMMs, as it restricts their ability to develop a comprehensive understanding of image content. 

\begin{figure}
    \centering
    
    \includegraphics[width=\linewidth]{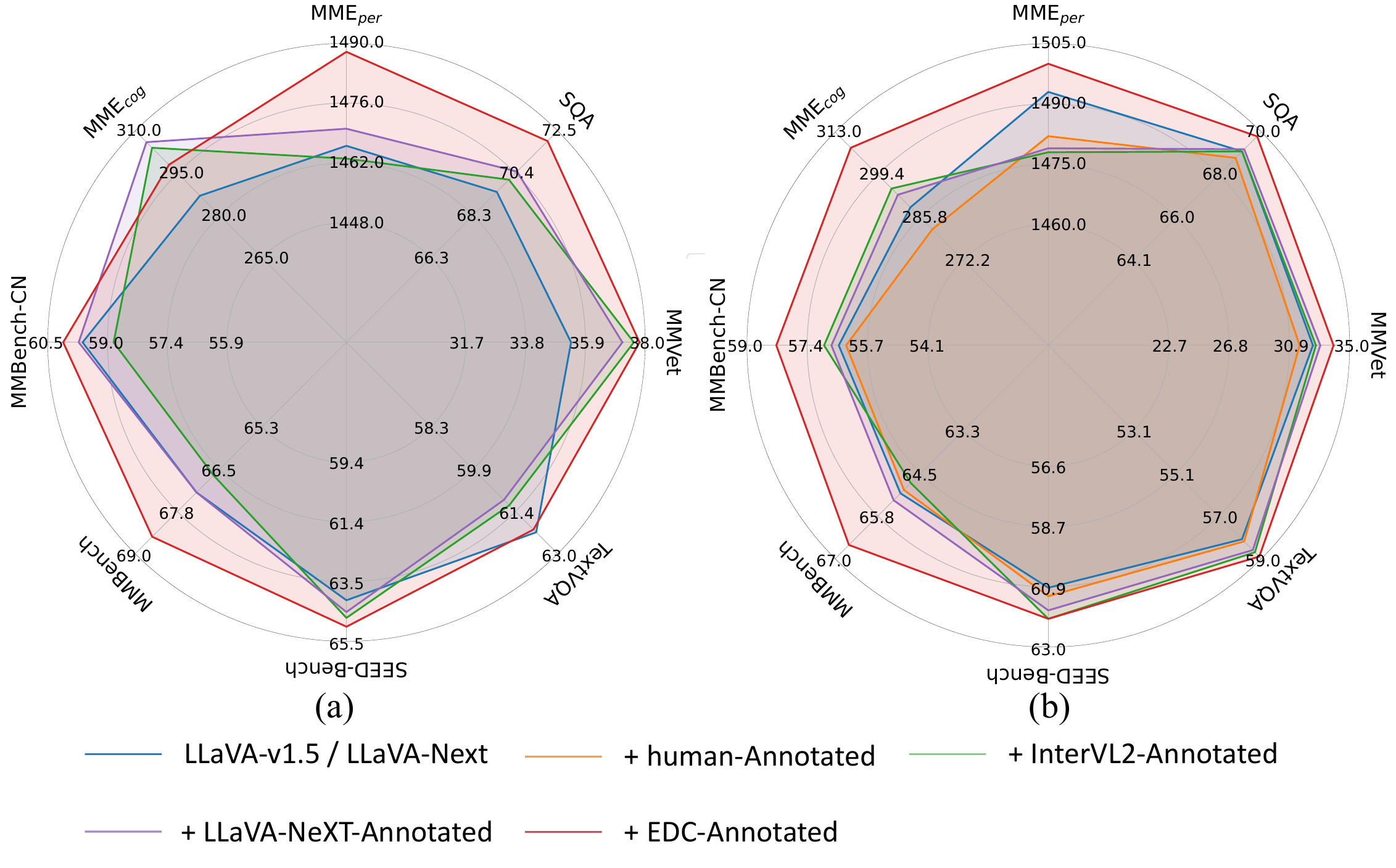}
    \vspace{-1.5em}
    \caption{Comparisons of caption quality. (a) and (b) show the downstream task performance of LLaVA-v1.5 and LLaVA-NeXT after pretraining with different image captions.}
    \label{fig:captionquality}
    \vspace{-1.5em}
\end{figure}

On the other hand, captions generated by SoTA LMMs~\cite{achiam2023gpt4,chen2024internvl}, while an improvement over human annotations, still exhibit notable deficiencies. As shown in Figure~\ref{fig:drawbacksofcurrentcaptions} (b), we selected 8 objects and 5 key attributes from the image to analyze and compare the captions generated by different methods. The key attributes include fine-grained attributes, spatial relation, and HOI. It is clear that captions generated by LMMs are more detailed than those annotated by humans. They not only describe more objects but also more attributes. However, captions from LMMs still much improvement space, as they tend to overlook important objects and attributes. For example, in Figure~\ref{fig:drawbacksofcurrentcaptions}(a), Object 6 is completely ignored by all LMMs. Furthermore, crucial 3D spatial relationships between objects are also missing, which is particularly problematic for tasks that require a comprehensive understanding of the scene's structure~\cite{li2023seed-bench}.

Toward this end, we propose an Enhancing Descriptive Captions Engine (EDC), designed to enable efficient and cost-effective image captioning. 
We leverage visual specialists~\cite{zhu2023quantized,chen2023group,stevens2024bioclip,yang2024depth} to replicate various human visual capabilities, and subsequently employ large language models (LLMs)~\cite{touvron2023llama2} to simulate the human cognitive process. This combined approach enables us to generate high-quality image captions by closely mimicking the way humans perceive and interpret visual information. Notably, EDC relies solely on open-source visual expert models and LLMs, significantly reducing annotation costs.

Specifically, we leverage existing visual specialists to obtain instance-level and relational attributes within images. Instance-level attributes focus on object low-level and fine-grained attributes (e.g., depth, emotion and fine-grained categories). Relational-level attributes capture interactions and relationships between objects (e.g., relative location and HoI). Next, we use prompts to guide LLMs in combining object attributes into region captions. Finally, prompts are used again to integrate these region captions with relational-level attributes, producing a comprehensive and detailed image caption. Since EDC utilizes multiple off-the-shelf visual specialists, the resulting captions capture a wide array of detailed attributes and nuanced relationships, leading to richer and more precise image descriptions. As shown in Figure~\ref{fig:drawbacksofcurrentcaptions}(b), captions annotated by EDC contain the most comprehensive information among all methods. Figure~\ref{fig:captionquality} quantitatively demonstrates that the captions generated by EDC provide greatest benefit to LMMs.

We applied our EDC to annotate a large-scale dataset of 1.1 million images, consisting of 1 million diverse images (EDC-1M) and 118K real-world scene images (EDC-118K). Experiments were conducted using both LLaVA-v1.5 and LLaVA-NeXT models. The results show that the highly detailed captions generated by EDC significantly enhance the perceptual capabilities of large multimodal models (LMMs), improving visual-language alignment. The outstanding performance of both LLaVA-v1.5 and LLaVA-NeXT across 14 benchmarks further underscores the effectiveness of our approach, confirming that the generated image captions are of exceptional detail and quality.




\begin{figure*}
    \centering
    \vspace{-.5em}
    \includegraphics[width=\linewidth]{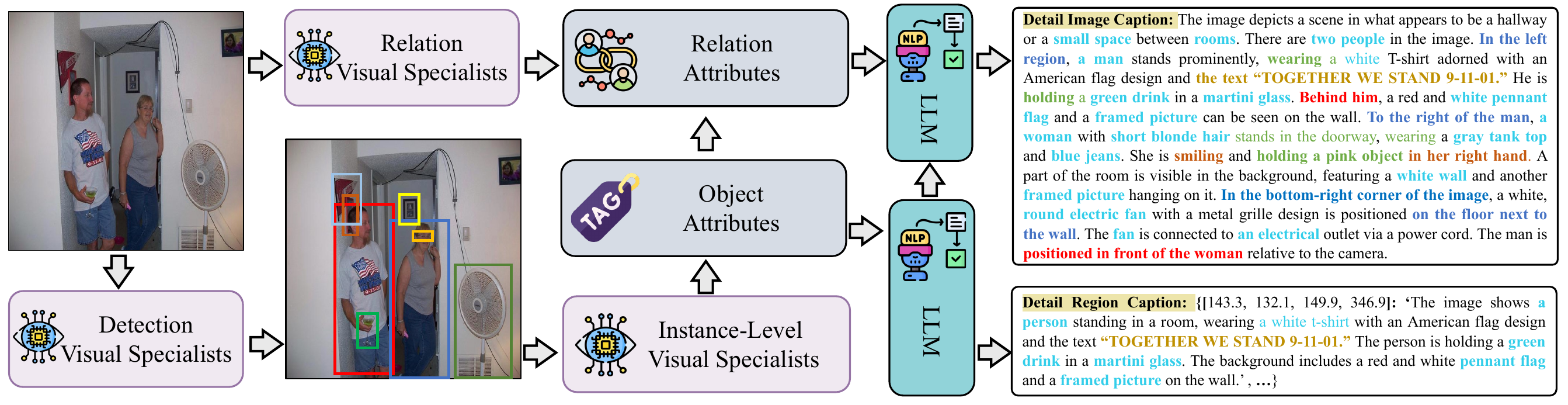}
    \vspace{-1.8em}
    \caption{The EDC pipeline first utilizes various visual specialists to extract both Object and Relation attributes. Then, it uses an LLM to integrate the object attributes into detailed region captions, followed by combining the region captions with relational attributes to generate a comprehensive image caption.}
    \label{fig:farmework}
    \vspace{-2.em}
\end{figure*}

\section{Related Work}

\subsection{Large multi-modality models}

Recently, there have been significant advancements in the development of large multi-modality models (LMMs), particularly in aligning and integrating pretrained vision encoders with pretrained language models to enhance their reasoning and perception capabilities~\cite{wang2023cogvlm,ye2023mplug2,sun2024improving}. A key line of research focuses on designing effective bridging mechanisms between these two modalities. For instance, Flamingo~\cite{alayrac2022flamingo} inserts new gated cross-attention layers between existing pretrained and frozen LM layers, that bridges powerful pretrained vision-only and language-only models. BLIP-2~\cite{li2022blip} employs a lightweight querying transformer that facilitates interaction between vision and language representations with minimal additional computational overhead. Similarly, Qwen-VL~\cite{bai2023qwen} adopts an approach akin to the querying transformer, while LLaVA~\cite{llava} utilizes a simple projection module to map visual embeddings to the language model’s input space. The Llama3 Herd of Models~\cite{gao2023llamaadapter2} extends this paradigm by incorporating an adapter mechanism akin to those in LLaVA~\cite{llava} and BLIP-2~\cite{li2022blip}, further optimizing the connection between visual and language components. In contrast, Some works~\cite{zhang2024vision} adopts a token-based fusion strategy, directly concatenating multimodal tokens—including both image and text tokens—before processing them within a unified transformer architecture. Emu~\cite{sun2023emu}, on the other hand, employs a causal transformer to transform image encodings into discrete token representations, improving their compatibility with autoregressive language models.

Beyond bridging mechanisms, another research has focused on improving the efficiency and effectiveness of visual encoding architectures. One promising approach involves partitioning images into large blocks, allowing for the extraction of fine-grained features while maintaining computational efficiency. Notable works leveraging this strategy include Monkey~\cite{li2024monkey}, Qwen2-VL~\cite{wang2024qwen2-vl}, and LLaVA-NeXT~\cite{li2024llavanext-strong}, all of which aim to enhance the granularity of visual representations while optimizing model efficiency.

In addition to architectural innovations, significant efforts have been made to improve LMMs through large-scale high-quality pretraining and fine-tuning data. High-quality and diverse datasets play a crucial role in shaping the capabilities of these models, as evidenced by works such as Qwen-VL~\cite{bai2023qwen}, Kosmos~\cite{peng2023kosmos}, Otter~\cite{li2023otter}, Shikra~\cite{chen2023shikra}, MiniGPT~\cite{zhu2023minigpt4}, and CogVLM~\cite{wang2023cogvlm}. Despite the impressive progress enabled by these models, relatively few studies have systematically explored the large-scale collection and curation of high-quality image-text datasets—an essential factor in further advancing LMM capabilities. The development of improved data collection pipelines, including strategies for filtering, augmentation, and quality assessment, remains a critical open research direction for the field.

\begin{table*}[ht]
\centering
\setlength\tabcolsep{4pt}
\renewcommand\arraystretch{1.02}
\setlength{\tabcolsep}{3.8mm}{
\caption{Summary of attributes our approach extracts through visual specialists. It includes the specific attribute names, the models used, and the extraction process for each.}
\resizebox{\textwidth}{!}{%
\begin{tabular}{l|l|l}
\hline
 \textbf{Attributes} & \textbf{Visual Specialists} & \textbf{Detailed Process} \\ \hline
\rowcolor{green!8} \multicolumn{3}{l}{\textit{Instance-Level}} \\ \hline
Size & Detection model & Using \textcolor{red}{\textbf{the area of the bounding box}} to measure the size of the instance. \\ \arrayrulecolor{lightgray} \hline \arrayrulecolor{black} 
 Depth & Depth \& Detection model & \textcolor{red}{\textbf{Average the depth map values within the bounding box region}} to obtain the depth information. \\ \arrayrulecolor{lightgray} \hline \arrayrulecolor{black} 
 Emotion & Emotion model & If the detected region is labeled as \textcolor{red}{\textbf{"person"}}, an emotion model is used \textcolor{red}{\textbf{to extract an emotion label}}. \\ \arrayrulecolor{lightgray} \hline \arrayrulecolor{black} 

 OCR & OCR Model & Using an OCR model to \textcolor{red}{\textbf{extract the text content and bounding box}} from the region. \\ \arrayrulecolor{lightgray} \hline \arrayrulecolor{black} 
 Animal & \multirow{7}{*}{Fine Grained model} & A fine-grained recognition model to identify \textcolor{red}{\textbf{specific species of the animal}}. \\ \arrayrulecolor{lightgray} \cline{1-1} \cline{3-3} \arrayrulecolor{black} 
 Plants &  & A fine-grained recognition model to identify \textcolor{red}{\textbf{specific species of the plants}}. \\ \arrayrulecolor{lightgray} \cline{1-1} \cline{3-3} \arrayrulecolor{black} 
 Aircrafts &  & A fine-grained recognition model to identify \textcolor{red}{\textbf{specific model of the aircraft}}. \\ \arrayrulecolor{lightgray} \cline{1-1} \cline{3-3} \arrayrulecolor{black} 
 Logo &  & A fine-grained recognition model to \textcolor{red}{\textbf{identify logos}} in the region. \\ \arrayrulecolor{lightgray} \cline{1-1} \cline{3-3} \arrayrulecolor{black} 
 Landmark &  & A fine-grained recognition model to \textcolor{red}{\textbf{identify landmarks}} within the region. \\ \arrayrulecolor{lightgray} \cline{1-1} \cline{3-3} \arrayrulecolor{black} 
 Food &  & A fine-grained recognition model to identify \textcolor{red}{\textbf{specific species of the food}}. \\ \arrayrulecolor{lightgray} \cline{1-1} \cline{3-3} \arrayrulecolor{black}
 Celebrity &  & Using a fine-grained recognition model to \textcolor{red}{\textbf{identify celebrity}} within the region. \\ \hline
 \rowcolor{green!8} \multicolumn{3}{l}{\textit{Relation}} \\ \hline 
 P2O relation & HOI Model & \begin{tabular}[c]{@{}l@{}}Using an HOI model to \textcolor{red}{\textbf{determine the relationship between the person and the object}}, while \\ the bounding boxes of both the person and the object define their respective regions.\end{tabular} \\ \arrayrulecolor{lightgray} \hline \arrayrulecolor{black}
Count & Detection model & \textcolor{red}{\textbf{Counting the number of all objects}} in the image based on the detection results. \\ \arrayrulecolor{lightgray} \hline \arrayrulecolor{black}
2D Absolute Location & Detection model & \begin{tabular}[c]{@{}l@{}}Using the bounding box to \textcolor{red}{\textbf{determine the instance's position within the image}}, including regions  \\ such as \textbf{left}, \textbf{right}, \textbf{top}, \textbf{bottom}, \textbf{center}, \textbf{top-left}, \textbf{bottom-left}, \textbf{top-right}, and \textbf{bottom-right}.\end{tabular} \\ \arrayrulecolor{lightgray} \hline \arrayrulecolor{black} 

2D Relative Location & Detection model & \begin{tabular}[c]{@{}l@{}}Using the bounding box to \textcolor{red}{\textbf{determine the  relative position among multiple objects within}} \\ \textcolor{red}{\textbf{the image}},  including regions such as \textbf{left}, \textbf{right}, \textbf{near}, \textbf{next to},  \textbf{close by}, and so on.\end{tabular} \\ \arrayrulecolor{lightgray} \hline \arrayrulecolor{black}
3D Relative Location & Detection \& Depth model & \begin{tabular}[c]{@{}l@{}l@{}}Using the depth attributes of different instances to \textcolor{red}{\textbf{capture the 3D spatial relationships of objects}} \\ \textcolor{red}{\textbf{relative to the camera}}, such as "Instance\_A is \textbf{in front of} Instance\_B" or "Instance\_A is \textbf{behind of} \\ Instance\_B" relative to the camera.\end{tabular} \\ \arrayrulecolor{lightgray} \hline \arrayrulecolor{black} 

 \hline
\end{tabular}

\label{tab:attribute_list}
}}
\vspace{-1.8em}
\end{table*}

\subsection{Descriptive captions}

High-quality descriptive captions play a crucial role in improving the performance of LMMs by providing detailed and context-rich textual descriptions of visual content~\cite{chen2023sharegpt4v,krishna2017VG}. Several large-scale datasets have been developed to facilitate research in image captioning and vision-language modeling~\cite{stefanini2022show,tu2024smart,guo2024unk}. One of the most widely used datasets, Conceptual Captions 3M (CC3M)~\cite{cc3m}, is constructed by harvesting image-alt text pairs from web data. To ensure caption quality, an automatic pipeline is employed to extract, filter, and transform candidate image-caption pairs, improving both linguistic and semantic relevance. Building upon CC3M, Conceptual Captions 12M (CC12M)~\cite{cc12m} extends this methodology with a focus on enhancing long-tail visual recognition. CC12M significantly expands the dataset’s diversity by incorporating a broader range of objects, scenes, and rare visual concepts, thereby improving the representation of real-world imagery. In addition to automatically curated datasets, manually annotated caption datasets have been developed to ensure greater accuracy and contextual relevance. SBU Captions\cite{ordonez2011sbucap} is a large-scale dataset sourced from Flickr~\cite{flickr30k}, where user-written descriptions provide natural language captions for images. These captions capture a diverse range of visual content, including both common and highly specific scenes. Similarly, COCO-Captions\cite{chen2015cococap} extends the well-known COCO dataset by incorporating human-annotated image descriptions collected through Amazon Mechanical Turk (AMT). While manually curated captions tend to be more accurate and aligned with human perception, they often lack finer-grained details and contextual depth, limiting their usefulness for training models that require richer semantic understanding.

Toward this end, recent methods have leveraged LMMs to generate more detailed image captions automatically. ShareGPT4V\cite{chen2023sharegpt4v} and DenseFusion\cite{Densefusion} represent notable examples of this approach, utilizing large-scale vision-language models to generate captions with enhanced descriptive richness. These methods rely on advanced language models such as GPT-4V\cite{chen2023sharegpt4v} to enrich captions with finer details that may be overlooked by human annotators. However, LMM-based caption generation still faces notable challenges, particularly regarding the accuracy and reliability of the generated descriptions. Issues such as over-simplification, hallucination, and misinterpretation of image content remain prevalent, potentially reducing caption quality and trustworthiness\cite{image_Tex,li2023seed-bench}. In this paper, we propose EDC pipeline, which integrates a diverse set of visual expert models specialized in different aspects of image understanding, including object recognition, scene parsing, and fine-grained attribute detection. By leveraging multiple visual experts, our method enhances the accuracy and contextual relevance of generated captions, reducing hallucinations common in LMM-based captioning. This approach achieves a better balance between detail richness and factual correctness, advancing the state-of-the-art in descriptive caption generation.




\section{Approach}
As shown in Figure~\ref{fig:farmework}, EDC leverage existing visual specialists
to extract visual properties
for improving descriptive captions.
We explore two kinds of properties:
object-level attributes
and object-relations.
Our approach consists of:
object localization using a SoTA open-world object detection specialist,
object property extraction 
using various specialists, 
and object relation extraction between objects.


\subsection{Object Attributes}

\noindent\textbf{Object localization.}
We combine in-domain detection models~\cite{chen2023group,meng2021conditional} and open-world detection models~\cite{joseph2021towards}
for robust detection,
merging bounding boxes from both models with confidence scores above 0.5. 
The object detection model outputs the location and semantic information of objects that existed in the image.
After detecting these regions, we apply Non-Maximum Suppression (NMS) to eliminate redundant or overlapping boxes. The IoU threshold for NMS is $0.75$.


\noindent\textbf{Attribute extraction by visual specialists.}
Table~\ref{tab:attribute_list} presents all the instance-level attributes involved in EDC, along with their extraction processes.
The attributes currently include three main parts.
(1) Fine-grained object category attributes.
This is rarely explored in current multimodal models~\cite{chen2023sharegpt4v}.
 In EDC, we incorporate fine-grained details by introducing various specialized models, covering categories such as \textit{animals, plants, food, logo, aircraft, landmarks, and celebrities}.
The animal and plant attributes contain the fine-grained category of $891k$ and $427k$ species of animals and plants, respectively. Each species within this category possesses unique characteristics and behaviors, making it a rich and varied classification. The food attributes include a variety of food types commonly found in daily life, while logos are graphic symbols that serve as visual identifiers for conveying messages. Landmarks are the famous tourist attractions that hold cultural, historical, or geographical significance. Aircraft refers to the machines designed for flight, including airplanes, helicopters, drones, and other aerial vehicles. Celebrities are individuals widely recognized in public life, entertainment, sports, or other fields. All this specific and fine-grained information could be regarded as the external world knowledge, aligning the textual content in the image with basic human cognition.
(2) Low-level and emotion attributes.
It includes \textit{emotion}, \textit{depth}, and \textit{size}.
(3) OCR.
It is one of the important attributes
for multimodal models.
The specific visual expert models we use will be presented in the \textbf{Appendix A.1}.


\subsection{Object Relation}

EDC can extract relationships between multiple objects. 
We consider three categories: the interactions between humans and objects, 
the 2D as well as 3D relative positional relationships among different objects, 
and the object counting information. 
Table~\ref{tab:attribute_list} presents the relation attributes and their extraction process.

The human-object interaction provides essential information about the actions and activities performed by humans with the objects.
We utilize the human-object interaction (HOI) model to detect interactive activities between humans and objects in the image. The interactions detected by the HOI model can be used to supplement events not mentioned in the caption. 

\begin{table}[]\footnotesize
    \caption{Human evaluation of attribute richness, conducted on 100 validation samples with 10 volunteers.}
    \vspace{-1.em}
 \label{tab:human_eval}
 \centering
 \renewcommand\arraystretch{1.2}
 \setlength{\tabcolsep}{4.5mm}{
 \resizebox{1.0\linewidth}{!}
{
\begin{tabular}{c|c|c|c}
\hline
\textbf{Attributes} & \textbf{InternVL2} & \textbf{LLaVA-NeXT} & \textbf{EDC} \\ \hline
\textbf{Spatial Relation} & 0.57 & 0.62 & \textcolor{red}{0.75} \\ 
\textbf{HOI} & 0.92 & 0.86 & \textcolor{red}{0.92} \\ 
\textbf{Fine-Grained} & 0.16 & 0.08 & \textcolor{red}{0.24} \\ 
\textbf{OCR} & 0.26 & 0.33 & \textcolor{red}{0.48} \\ 
\textbf{Emotion} & 0.23 & 0.14 & \textcolor{red}{0.47} \\ 
\textbf{Location} & 0.36 & 0.59 & \textcolor{red}{0.81} \\ \hline
\end{tabular}}}
\vspace{-2.em}
\end{table}


The 2D positional information captures the spatial relationships of objects, comprising both 2D Absolute Location and 2D Relative Location. The 2D Absolute Location describes an object’s position relative to the image (e.g., Object A is on the left side of the image), while the 2D Relative Location describes positional relationships between objects (e.g., Object A is next to Object B). We use the bounding boxes of objects to determine their positional relationships.

The 3D relative positional information captures the spatial relationships of objects in 3D space, defining both their absolute positions in the scene and their relative positioning (e.g., Object A is in front of Object B or at a specific angle). This information enhances the understanding of a scene's 3D structure and provides richer spatial awareness. We leverage the depth differences between objects to determine their 3D positional relationships.

\subsection{Captioning with Attributes}

\noindent\textbf{Region captioning.}
We use a large language model (Qwen2-72B-AWQ) 
to integrate the object attributes
with the caption from a large multimodal model
(InternVL2-26B). We design a set of structured prompts that guide LLM to generate fine-grained descriptions by integrating visual details such as category, color, texture, spatial location, and human-object interactions into coherent natural language.
For example, when the fine-grained model for \textit{``animals''} predicts a specific class label like \textit{$\{animal\_name\}$}, we use the following prompt to guide the LLM in incorporating it into the caption:
\vspace{-.5em}
\begin{tcolorbox}[colback=green!5, colframe=black!10]
\textit{``{$\{cat\_name\}$} exists in the region and {$\{animal\_name\}$} is a subclass of {$\{cat\_name\}$}; use {$\{animal\_name\}$} in the caption; otherwise, do not mention \textit{$\{animal\_name\}$}''}
\end{tcolorbox}
\vspace{-.5em}
Here, \textit{$\{cat\_name\}$} denotes the coarse-grained label predicted by the detection model. The complete prompt engineering strategy is detailed in the \textbf{Appendix A.3}.



\begin{table*}[!t]
\centering
\setlength\tabcolsep{4pt}
\renewcommand\arraystretch{1.05}
\setlength{\tabcolsep}{8.4mm}{
\caption{Comparison of Different Image Captioning Annotation Methods. The \textcolor{red}{red} and \textcolor{blue}{blue} colors respectively represent the optimal and suboptimal results on each benchmark.}
\vspace{-1.em}
\label{tab:campare_caption}
\resizebox{\textwidth}{!}{%
\begin{tabular}{l|cccccc}
\hline
\textbf{Annotation Methods}    & \textbf{GQA}  & \textbf{ScienceQA} & \textbf{MMBench} & \textbf{MM-Vet} & \textbf{SEED$^I$} & \textbf{SEED-Bench} \\ \hline
\rowcolor{green!8}\multicolumn{7}{l}{\textit{LLaVA-v1.5-7B }}           \\ \hline
+ShareGPT4V~\cite{chen2023sharegpt4v}   & 63.3 & 68.4      & \textcolor{blue}{\textbf{68.8}}    & 37.6   & 69.7  & 61.9       \\ 
+DenseFusion~\cite{Densefusion} & \textcolor{blue}{\textbf{64.0}} & 69.3      & \textcolor{red}{\textbf{69.2}}    & \textcolor{blue}{\textbf{37.8}}   & \textcolor{blue}{\textbf{70.1}}  & \textcolor{blue}{\textbf{62.3}}       \\ 
+DCI~\cite{DCI}          & 63.1 & 69.2      & 64.3    & 30.2   & 66.7  & 61.2       \\ 
+DOCCI~\cite{DOCCI}        & 63.2 & \textcolor{blue}{\textbf{69.4}}      & 65.0    & 32.3   & 66.6  & 61.1       \\ 
+IT~\cite{image_Tex}          & 62.9 & 68.0      & 65.5    & 33.1   & 67.2  & 61.7       \\ 
+EDC          & \textcolor{red}{\textbf{64.2}} & \textcolor{red}{\textbf{71.0}}      & \textcolor{red}{\textbf{69.2}}    & \textcolor{red}{\textbf{38.2}}   & \textcolor{red}{\textbf{70.3}}  & \textcolor{red}{\textbf{64.3}}       \\ \hline
\rowcolor{green!8}\multicolumn{7}{l}{\textit{LLaVA-NeXT-7B}}            \\ \hline
+Recap~\cite{coco-llava}        & \textcolor{blue}{\textbf{65.0}} & 70.5      & 67.2    & 37.2   & -     & 64.5       \\ 
+DCI~\cite{DCI}          & 64.7 & 70.7      & 68.0    & 38.0   & 70.6  & 64.1       \\ 
+DOCCI~\cite{DOCCI}        & 64.8 & 71.1      & 68.5    & 36.7   & 70.8  & 64.5       \\ 
+IT~\cite{image_Tex}           & 64.9 & \textcolor{red}{\textbf{71.3}}      & \textcolor{blue}{\textbf{68.6}}    & \textcolor{blue}{\textbf{38.1}}   & \textcolor{blue}{\textbf{71.4}} & \textcolor{blue}{\textbf{65.4}}       \\ 
+EDC          & \textcolor{red}{\textbf{65.2}} & \textcolor{blue}{\textbf{71.2}}      & \textcolor{red}{\textbf{69.3}}    & \textcolor{red}{\textbf{40.1}}   & \textcolor{red}{\textbf{72.2}}  & \textcolor{red}{\textbf{65.7}}       \\ \hline
\end{tabular}}}
\vspace{-2.em}
\end{table*}

\noindent\textbf{Image captioning.}
We combine object relation attributes, region location information (for object grounding), and reference captions to generate improved image captions. We use a LLM to integrate object-level attributes through prompt engineering. For example, to incorporate the 3D spatial relationship between two objects, we design prompts such as:
\vspace{-.5em}
\begin{tcolorbox}[colback=green!5, colframe=black!10]
\textit{``Relative to the camera, the $\{category\_0\}$ in $\{bbox\_0\}$ is $\{3d\_relation\}$ the $\{category\_1\}$ in $\{bbox\_1\}$.''}
\end{tcolorbox}
\vspace{-.5em}
Here, $\{3d\_relation\}$ captures spatial cues such as \textit{``in front of''} or \textit{``behind of''}, which are computed based on the depth difference between the two objects. We use this mechanism to update the image caption with 3D relative localization information. The complete set of prompt templates is provided in the \textbf{Appendix A.3}.

\subsection{Analysis}

\noindent\textbf{Dataset Description.}
We leverages EDC to generate more detailed annotations for publicly available image datasets, resulting in two enhanced datasets: (1) EDC-1M, which provides dense annotations for 1 million diverse images from Densefusion~\cite{Densefusion}, covering a wide range of objects and scenes, and (2) EDC-118K, which includes refined annotations for 118,000 complex scene images from the COCO dataset~\cite{lin2014coco}. We provide a detailed analysis of the EDC-annotated image captions in the \textbf{Appendix B}.

\noindent\textbf{Attribute richness.}
We randomly select 100 images from the EDC-118K and invite five independent evaluators 
to assess the captions for each image. 
The evaluators analyzed and recorded the occurrence of 
the attributes 
within the captions, such as spatial relations, HOI, and OCR (optical character recognition) details.

The results as show in Table~\ref{tab:human_eval}. Compared to other models and human annotations, our approach demonstrated a greater ability to capture and express a wider diversity of visual attributes present in the images. This suggests that our approach provided richer and more detailed descriptions of the visual content.

\noindent\textbf{Comparison.}
We conducted a comparison of human annotations, internVL2-26B, LLaVA-NeXT-34B, and EDC-annotated 118K datasets on both LLaVA-v1.5 and LLaVA-NeXT. The results as shown in Figure~\ref{fig:captionquality}, the captions annotated by EDC significantly improve the performance of LLaVA-v1.5 and LLaVA-NeXT. Compared to human and generalist multimodal model annotations, EDC-annotated captions are richer and more detailed, offering deeper context and capturing finer nuances. This enhanced descriptive quality leads to better performance in downstream tasks.

\section{Experiments}

\begin{table*}[!t]
\centering
\setlength\tabcolsep{4pt}
\renewcommand\arraystretch{1.05}
\setlength{\tabcolsep}{5.2mm}{
\caption{Performance on seven General Visual Question Answering benchmarks. The \textcolor{red}{red} and \textcolor{blue}{blue} colors respectively represent the optimal and suboptimal results on each benchmark. $*$ indicates the use of LLaVA-NeXT’s open-source SFT data, with certain private data excluded.}
\vspace{-1.em}
\label{tab:vqa}
\resizebox{\textwidth}{!}{%
\begin{tabular}{l|l|ccccccc}
\hline
\multirow{2}{*}{\textbf{Model}} & \multirow{2}{*}{\textbf{LLM}} & \multicolumn{7}{c}{\textbf{Visual Question Answering Benchmarks}} \\ \cline{3-9}
   &  & \textbf{VQAv2} & \textbf{DocVQA} & \textbf{OKVQA} & \textbf{GQA} & \textbf{TextVQA} & \textbf{ScienceQA} & \textbf{Ai2d} \\ \hline
 \rowcolor{green!8}\multicolumn{9}{l}{\textit{Low Resolution Models}} \\ \hline
 BLIP2~\cite{li2022blip} & Flan-T5 & 41.0 & - & 45.9 & 41.0 & 42.5 & 61.0 & - \\
 InstructBLIP~\cite{dai2024instructblip} & Vicuna-7B & - & - & - & 49.2 & 50.1 & 60.5 & 40.6 \\
 InstructBLIP~\cite{dai2024instructblip} & Vicuna-13B & - & - & - & 49.5 & 50.7 & 63.1 & - \\
 IDEFICS-Instruct~\cite{laurenccon2024obelics} & LLaMA-65B & 60.0 & - & 36.9 & - & 32.9 & 61.8 & 54.8 \\
 OpenFlamingo~\cite{openflamingo} & MPT-7B & 53.0 & - & 38.3 & - & 28.3 & 44.8 & - \\
 InternVL-Chat~\cite{chen2024internvl} & Vicuna-7B & 79.3 & - & 51.8 & 62.9 & 57.0 & - & - \\
 Qwen-VL-Chat~\cite{bai2023qwen} & Qwen-7B & 78.2 & \textcolor{red}{\textbf{62.6}} & 56.6 & 57.5 & \textcolor{red}{\textbf{61.5}} & 68.2 & - \\
 mPLUG-Owl2~\cite{ye2023mplug2} & LLaMA-7B & 79.4 & - & \textcolor{red}{\textbf{57.7}} & 56.1 & 58.2 & \textcolor{blue}{\textbf{68.7}} & \textcolor{blue}{\textbf{55.7}} \\
 LLaVA-v1.5~\cite{liu2023llava1.5} & Vicuna-7B & 78.5 & 28.1 & - & 62.0 & 58.2 & 66.8 & 55.5 \\
 ShareGPT4V~\cite{chen2023sharegpt4v} & Vicuna-7B & 80.6 & - & - & \textcolor{blue}{\textbf{63.3}} & 60.4 & 68.4 & - \\
 \rowcolor{gray!15} LLaVA-v1.5(Ours) & Vicuna-7B & \textcolor{red}{\textbf{80.9}} & \textcolor{blue}{\textbf{39.1}}  & \textcolor{blue}{\textbf{57.2}} & \textcolor{red}{\textbf{64.2}} & \textcolor{blue}{\textbf{61.4}} & \textcolor{red}{\textbf{71.0}} & \textcolor{red}{\textbf{59.4}} \\ \hline
 \rowcolor{green!8}\multicolumn{9}{l}{\textit{High Resolution Models}} \\ \hline
 Monkey~\cite{li2024monkey} & Qwen-7B & 80.3 & 66.5 & \textcolor{red}{\textbf{61.3}} & 60.7 & \textcolor{red}{\textbf{67.6}} & 69.4 & 62.6 \\
 LLaVA-NeXT~\cite{li2024llavanext-strong} & Vicuna-7B & 81.8 & 74.4 & 54.3 & 64.2 & 64.9 & 70.1 & 66.6 \\
 LLaVA-NeXT+ReCap~\cite{coco-llava} & Vicuna-7B & - & \textcolor{blue}{\textbf{75.3}} & 44.3 & \textcolor{blue}{\textbf{65.0}} & - & \textcolor{blue}{\textbf{71.0}} & \textcolor{blue}{\textbf{66.9}} \\
 LLaVA-S$^2$~\cite{shi2025llavas2} & Vicuna-7B & 79.7 & - & - & 63.3 & 60.8 & 68.2 & - \\
 LLaVA-HR~\cite{luo2024feast} & Vicuna-7B & \textcolor{blue}{\textbf{81.9}} & - & \textcolor{blue}{\textbf{58.9}} & 64.2 & \textcolor{blue}{\textbf{67.1}} & 65.1 & - \\ \arrayrulecolor{lightgray} \hline \arrayrulecolor{black}
 \rowcolor{gray!15} LLaVA-NeXT$^*$(Ours) & Vicuna-7B & \textcolor{red}{\textbf{82.4}} & \textcolor{red}{\textbf{78.8}} & 57.2 & \textcolor{red}{\textbf{65.2}} & 64.8 & \textcolor{red}{\textbf{71.2}} & \textcolor{red}{\textbf{71.2}} \\ \hline
\end{tabular}
}}
\vspace{-1.em}
\end{table*}

In this section, we first present the implementation details, then compare the EDC-1M-trained model with state-of-the-art LMMs across multiple benchmarks. Finally, we perform ablation studies on the EDC-118K-trained model to validate the effectiveness of EDC.

\begin{table*}[!t]
\centering
\setlength\tabcolsep{4pt}
\renewcommand\arraystretch{1.05}
\setlength{\tabcolsep}{2mm}{

\caption{Performance on seven Large Multi-Modal benchmarks. The \textcolor{red}{red} and \textcolor{blue}{blue} colors respectively represent the optimal and suboptimal results on each benchmark. $*$ indicates the use of LLaVA-NeXT’s open-source SFT data, with certain private data excluded.}
\vspace{-1.em}
\label{tab:lvlm}
\resizebox{\textwidth}{!}{%
\begin{tabular}{l|c|c|ccccccc}
\hline
 \textbf{Method} & \textbf{Vision Encoder} & \textbf{Language Model} & \textbf{MMBench-CN} & \textbf{MMBench} & \textbf{MM-Vet} & \textbf{SEED$^I$} & \textbf{SEED-Bench} & \textbf{MMMU} & \textbf{POPE} \\ \hline
 \rowcolor{green!8}\multicolumn{10}{l}{\textit{Low Resolution Models}} \\ \hline
  BLIP-2~\cite{li2022blip} & ViT-g (1.3B) & Vicuna-7B & - & - & 22.4 & 46.4 & - & - & 85.3 \\
  MiniGPT-4~\cite{zhu2023minigpt4} & ViT-g (1.3B) & Vicuna-7B & 11.9 & 23.0 & 22.1 & - & 47.4 & 23.6 & - \\
  InstructBLIP & ViT-g (1.3B) & Vicuna-7B & 23.7 & 36.0 & 26.2 & 53.4 & - & 30.6 & 78.9 \\
 LLaMA-Adapter-v2~\cite{gao2023llamaadapter2} & ViT-L (0.3B) & LLaMA-7B & - & 39.5 & 31.4 & - & 32.7 & - & - \\
 OpenFlamingo~\cite{openflamingo}& ViT-L (0.3B) & MPT-7B & - & 5.7 & 24.8 & - & 42.7 & 26.3 & - \\
 Otter~\cite{li2023otter} & ViT-L (0.3B) & LLaMA-7B & - & 48.3 & 24.6 & - & 32.9 & - & - \\
 Qwen-VL-Chat~\cite{bai2023qwen} & ViT-G (1.9B) & Qwen-7B & 56.7 & 60.6 & - & 58.2 & - & 29.6 & - \\
 mPLUG-Owl2~\cite{ye2023mplug2} & ViT-L (0.3B) & Vicuna-7B & - & 64.5 & 36.2 & - & 57.8 & 34.7 & 86.2 \\
 LLaVA-v1.5~\cite{liu2023llava1.5} & ViT-L (0.3B) & Vicuna-7B & 57.6 & 64.3 & 30.5 & 66.2 & 58.6 & \textcolor{blue}{\textbf{35.3}} & \textcolor{blue}{\textbf{85.9}} \\
 ShareGPT4V~\cite{chen2023sharegpt4v} & ViT-L (0.3B) & Vicuna-7B & \textcolor{red}{\textbf{62.2}} & \textcolor{blue}{\textbf{68.8}} & \textcolor{blue}{\textbf{37.6}} & \textcolor{blue}{\textbf{69.7}} & \textcolor{blue}{\textbf{61.9}} & - & 85.7 \\
 \rowcolor{gray!15} LLaVA-v1.5(Ours) & ViT-L (0.3B) & Vicuna-7B & \textcolor{blue}{\textbf{60.0}} & \textcolor{red}{\textbf{69.2}} & \textcolor{red}{\textbf{38.2}} & \textcolor{red}{\textbf{70.3}} & \textcolor{red}{\textbf{64.3}} & \textcolor{red}{\textbf{36.3}} & \textcolor{red}{\textbf{86.4}} \\ \hline
 \rowcolor{green!8}\multicolumn{10}{l}{\textit{High Resolution Models}} \\ \hline
 LLaVA-NeXT~\cite{li2024llavanext-strong} & ViT-L (0.3B) & Vicuna-7B & \textcolor{blue}{\textbf{60.6}} & 67.4 & \textcolor{red}{\textbf{43.9}} & \textcolor{blue}{\textbf{70.2}} & \textcolor{blue}{\textbf{64.7}} & \textcolor{blue}{\textbf{35.1}} & 86.5 \\
  LLaVA-NeXT+ReCap~\cite{coco-llava} & ViT-L (0.3B) & Vicuna-7B & \textcolor{blue}{\textbf{-}} & 67.2 & - & - & 64.5 & - & 86.8 \\
 ShareGPT4V-S2~\cite{Densefusion} & ViT-L (0.3B) & Vicuna-7B & - & \textcolor{blue}{\textbf{68.0}} & 35.0 & 70.1 & 62.4 & - & 86.7 \\
 LLaVA-S$^2$~\cite{shi2025llavas2} & ViT-L (0.3B) & Vicuna-7B & - & 66.4 & 34.6 & 67.2 & 59.9 & - & 86.7 \\
 LLaVA-HR~\cite{luo2024feast} & ViT-L (0.3B) & Vicuna-7B & - & - & 31.2 & - & 64.2 & - & \textcolor{red}{\textbf{87.6}} \\ \arrayrulecolor{lightgray} \cline{1-10} \arrayrulecolor{black}
 \rowcolor{gray!15} LLaVA-NeXT$^*$(Ours) & ViT-L (0.3B) & Vicuna-7B & \textcolor{red}{\textbf{61.7}} & \textcolor{red}{\textbf{69.3}} & \textcolor{blue}{\textbf{40.1}} &\textcolor{red}{\textbf{72.2}}   & \textcolor{red}{\textbf{65.7}} & \textcolor{red}{\textbf{36.0}} &\textcolor{blue}{\textbf{87.0}} \\ \hline
\end{tabular}
}}
\vspace{-2.em}
\end{table*}

\subsection{implementation details}

\noindent\textbf{Model and Training Set.}  
We conduct experiments on LLaVA-v1.5~\cite{liu2023llava1.5} and LLaVA-NeXT~\cite{liu2024llavanext_15} to demonstrate the effectiveness of EDC. 
Specifically, we using CLIP-L~\cite{radford2021learning} as the visual encoder and Vicuna-v1.5~\cite{chiang2023vicuna} as the large language model. We adopt a two-stage training strategy: (1) Pre-Training Stage. We train only the projector for initial alignment. Then, following SharGPT4V~\cite{chen2023sharegpt4v}, we set the last 12 layers of the visual encoder in LLaVA-v1.5 as trainable and make the entire LLaVA-NeXT model trainable, following~\cite{coco-llava}, to further enhance perceptual capabilities. (2) Instruction Tuning Stage. We use the open-source LLaVA-mix-665K and LLaVA-NeXT-data to respectively train the LLaVA-v1.5 and LLaVA-NeXT models. The detailed training procedure is provided in the supplementary material.

\noindent\textbf{Evaluation Benchmarks.} 
We evaluate on seven visual question answering (VQA) tasks across domains such as document understanding, general knowledge, and scientific reasoning, including VQAv2~\cite{goyal2017vqav2}, DocVQA~\cite{mathew2021docvqa}, OKVQA~\cite{schwenk2022okvqa}, GQA~\cite{hudson2019gqa}, TextVQA~\cite{textvqa}, ScienceQA~\cite{scienceQA}, and Ai2d~\cite{ai2d}. Additionally, we evaluate performance on five widely used LMM benchmarks designed to test multimodal models on visual grounding, scene understanding, and generalization: MMBench~\cite{liu2023mmbench}, MM-Vet~\cite{yu2023mmvet}, SEED~\cite{li2023seed-bench}, MMMU~\cite{yue2024mmmu}, and POPE~\cite{li2023pope}.

\noindent\textbf{Comparison Method.} We compare LLaVA-v1.5 and LLaVA-NeXT models trained on EDC-1M against current SOTA MLLMs. Additionally, we evaluate the performance of various MLLMs in generating image captions, using the advanced models InternVL2-26B~\cite{chen2024internvl} and LLaVA-NeXT-34B~\cite{li2024llavanext-strong} to generate competitive captions on the EDC-118K images.

\subsection{Comparison with the State-of-the-art}
To validate the effectiveness of EDC captions, we compare them with several state-of-the-art image captioning methods on the LLaVA and LLaVA-Next benchmarks. These methods include ShareGPT4V~\cite{chen2023sharegpt4v}, DenseFusion~\cite{Densefusion}, ReCap~\cite{coco-llava}, DCI~\cite{DCI}, DOCCI~\cite{DOCCI}, and IT~\cite{image_Tex}. As shown in Table~\ref{tab:campare_caption}, models trained with EDC captions achieve better results on most evaluation metrics.

This demonstrates that EDC captions enhance LMM training by providing more comprehensive image descriptions, capturing finer-grained visual attributes and richer contextual information compared to existing methods. This performance gain can be attributed to the design of EDC, which integrates fine-grained object details, 3D spatial relations, and grounded region-level information via structured prompt engineering, offering more informative and aligned supervision signals.


\subsection{Main Results}

\noindent\textbf{VQA Benchmarks.} 
The results on six common visual question answering (VQA) datasets
are presented in Table~\ref{tab:vqa}. 
It is clear that LLaVA-v1.5 and LLaVA-NeXT, trained with EDC-1M, achieve state-of-the-art performance in both low-resolution and high-resolution settings. Compared to the baseline LLaVA-v1.5~\cite{liu2023llava1.5}, our model excels across all VQA benchmarks, demonstrating that high-quality image captions significantly enhance model performance. This highlights the crucial role of detailed and accurate captions in improving visual understanding for VQA tasks. Furthermore, when compared to the baseline LLaVA-NeXT~\cite{li2024llavanext-strong}, this improvement remains consistent, suggesting that the impact of high-quality captions is not dependent on the model variation. Additionally, compared to models like ShareGPT-4V, our model demonstrates superior performance across most VQA benchmarks. This improvement indicates that the captions generated by our EDC method provide richer and more comprehensive information. The model trained on EDC-1M demonstrates exceptional performance on datasets such as VQAv2~\cite{goyal2017vqav2} and Ai2D~\cite{ai2d}, highlighting that integrating detection models into EDC significantly enriches the diversity of objects in the generated captions, thereby boosting the model's performance on object recognition benchmarks. Furthermore, incorporating relational attributes into EDC enriches the captions with detailed inter-object relationships, enhancing the model's ability to capture and understand complex relationships, which further improves its performance on visual reasoning benchmarks such as GQA~\cite{hudson2019gqa}. However, performance on tasks like TextVQA~\cite{textvqa} is hindered by limitations in the open-source OCR model and the threshold settings; a high threshold restricts the model's ability to capture finer textual details.



\begin{table*}[!t]
\centering
\setlength\tabcolsep{4pt}
\renewcommand\arraystretch{1.}
\setlength{\tabcolsep}{6.4mm}{
\caption{Comparison of Different Image Captioning Annotation Methods.}
\vspace{-1.em}
\label{tab:ablation}
\resizebox{\textwidth}{!}{%
\begin{tabular}{l|ccccccc}
\hline
 \textbf{Annotation} & \textbf{OKVQA} & \textbf{GQA} & \textbf{ScienceQA} & \textbf{TextVQA} & \textbf{MMBench} & \textbf{MM-Vet} & \textbf{SEED-Bench} \\ \hline
 \rowcolor{green!8}\multicolumn{8}{l}{\textit{LLaVA-v1.5}} \\ \hline
 + human~\cite{coco} & 54.9 & 62.4 & 68.6 & 58.1 & 65.0 & 31.6 & 61.1 \\
  + InternVL2-26B~\cite{chen2024internvl} & 54.7 & \textcolor{blue}{63.0} & \textcolor{blue}{69.1} & 58.4 & 64.8 & 32.7 & \textcolor{blue}{61.8} \\
  + LLaVA-NeXT-34B~\cite{li2024llavanext-strong} & \textcolor{blue}{55.7} & 62.9 & 68.8 & \textcolor{blue}{58.7} & \textcolor{blue}{65.3} & \textcolor{blue}{33.0} & 61.7 \\ 
  + EDC & \textcolor{red}{\textbf{56.9}} & \textcolor{red}{\textbf{63.2}} & \textcolor{red}{\textbf{69.8}} & \textcolor{red}{\textbf{58.9}} & \textcolor{red}{\textbf{66.6}} & \textcolor{red}{\textbf{33.9}} & \textcolor{red}{\textbf{62.0}} \\\hline
  \rowcolor{green!8}\multicolumn{8}{l}{\textit{LLaVA-NeXT}} \\ \hline
 + InternVL2-26B~\cite{chen2024internvl} & 54.3 & \textcolor{blue}{65.1} & 70.1 & \textcolor{blue}{61.2} & 66.7 & \textcolor{blue}{37.3} & \textcolor{blue}{64.7} \\
  + LLaVA-NeXT-34B~\cite{li2024llavanext-strong} & \textcolor{blue}{54.3} & 65.0 & \textcolor{blue}{70.5} & 61.0 & \textcolor{blue}{67.2} & 37.2 & 64.5 \\ 
  + EDC & \textcolor{red}{\textbf{56.7}} & \textcolor{red}{\textbf{65.2}} & \textcolor{red}{\textbf{72.0}} & \textcolor{red}{\textbf{62.0}} & \textcolor{red}{\textbf{68.5}} & \textcolor{red}{\textbf{37.8}} & \textcolor{red}{\textbf{65.0}}\\\hline
\end{tabular}}}
\vspace{-1.em}
\end{table*}

\begin{figure*}
    \centering
    \includegraphics[width=\linewidth]{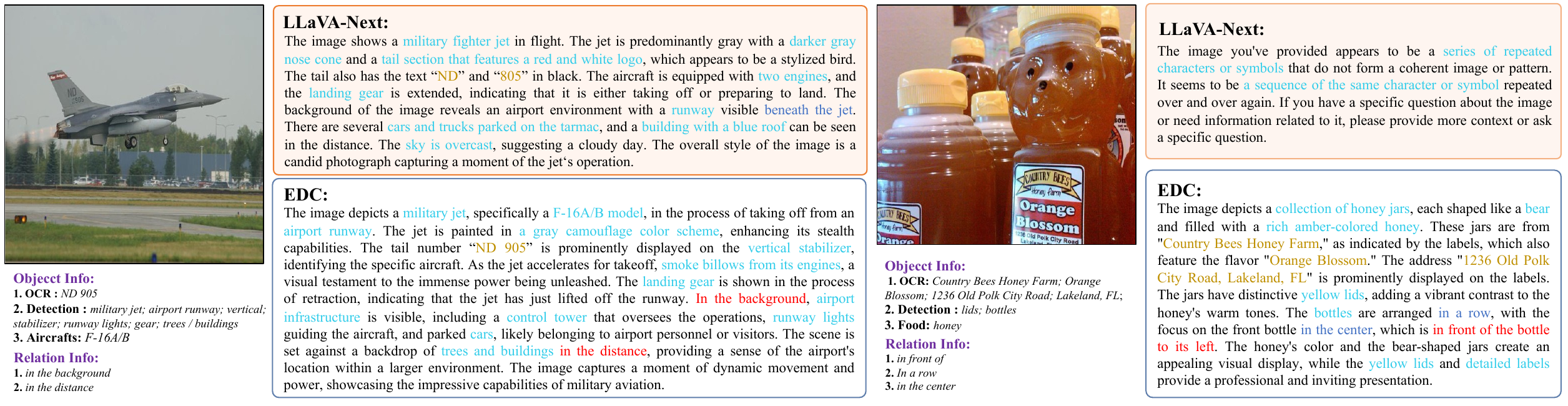}
    \vspace{-1.8em}
    \caption{Visualization of EDC's Attribute Fusion: Integrating Object and Relational Attributes for Richer, More Detailed Captions.}
    \label{fig:ablation_vis}
    \vspace{-2.em}
\end{figure*}

\noindent\textbf{Large Multi-Modal Benchmarks.} 
We further conduct the evaluation on five challenging large multi-modal benchmarks.
The experimental results are shown in Table~\ref{tab:lvlm}. It can be seen that both LLaVA-v1.5 and LLaVA-NeXT trained with EDC-1M achieve competitive performance on more complex LMM benchmarks, demonstrating that the improvements brought by EDC-1M are comprehensive. Our model outperforms both LLaVA-v1.5~\cite{liu2023llava1.5} and LLaVA-NeXT~\cite{liu2024llavanext_15} across all LMM benchmarks, demonstrating that high-quality image captions during pretraining significantly enhance model performance, even without altering the supervised fine-tuning (SFT) data. Compared to other image captioning methods, such as ShareGPT-4V~\cite{chen2023sharegpt4v}, EDC-generated captions provide richer and more comprehensive scene information, significantly boosting model performance across most LMM benchmarks. However, due to the lack of Chinese data in EDC-1M, the model performs poorly on MMBench-CN~\cite{liu2023mmbench}. This highlights the need for multilingual image captioning, which will be an area for future improvement in EDC. Additionally, the detection model in EDC introduces some noise, which may interfere with the model's ability to accurately capture objects, leading to decreased performance on tasks like POPE~\cite{li2023pope}. Therefore, reducing this noise will be a key focus for future improvements in EDC.





\subsection{Ablation Study}

\noindent\textbf{Comparing different annotation methods.} We compared different image annotation methods, including human annotations, GenerateList LMM annotations, and our EDC. Specifically, we annotated 118K COCO images and conducted comparisons on LLaVA-v1.5 and LLaVA-NeXT. The experimental results are shown in Table~\ref{tab:ablation}. We found that the image captions generated by EDC improve LMM performance on downstream tasks more effectively than other annotation methods. Notably, compared to captions annotated by internVL2, EDC's inclusion of object attributes significantly improves model performance on OKVQA and TextVQA tasks. Relational attributes enhance the model's understanding of multi-object relationships, leading to a notable increase in GQA performance. Furthermore, superior results on complex LMM benchmarks like MM-Vet, MMBench, and SEED highlight that EDC-generated captions provide rich and comprehensive scene information.

\noindent\textbf{Case Study.} 
Figure~\ref{fig:ablation_vis} presents example captions generated by the EDC engine and the general MLLM LLaVA-NeXT 34B.  It is evident that visual specialists within EDC capture detailed object and relational attributes, resulting in richer and more descriptive captions. For instance, in Figure~\ref{fig:ablation_vis}(a), the OCR information highlights the precision of OCR specialist model, while the fine-grained model’s identification of specific aircraft types further enhances caption informativeness. Additionally, the relational attributes significantly enrich description by providing detailed spatial relationships between objects, underscoring the advantages of EDC in capturing comprehensive scene details. More visualization results are provided in the \textbf{Appendix E}.

\vspace{-.5em}
\section{Conclusion}
\vspace{-.5em}
In this paper, we introduce EDC, a novel image captioning engine that leverages off-the-shelf visual specialists to enhance caption quality and detail. Unlike existing methods that rely solely on LMMs or human annotations, EDC integrates visual specialists to simulate human perception and LLMs to emulate cognitive reasoning. This dual approach enables the generation of captions that are both visually rich and contextually precise.  Extensive experiments show that incorporating visual specialists significantly improves model performance across various visual understanding and reasoning tasks, especially those requiring precise attribute and relationship recognition. Our work underscores the potential of specialized visual features in strengthening multimodal representations and offers a flexible framework for integrating additional visual expertise. We will release the EDC source code to support further research, allowing the community to seamlessly integrate additional visual specialists and enhance EDC’s captioning capabilities.

{
    \small
    \bibliographystyle{ieeenat_fullname}
    \bibliography{main}

\begin{thebibliography}{77}
\providecommand{\natexlab}[1]{#1}
\providecommand{\url}[1]{\texttt{#1}}
\expandafter\ifx\csname urlstyle\endcsname\relax
  \providecommand{\doi}[1]{doi: #1}\else
  \providecommand{\doi}{doi: \begingroup \urlstyle{rm}\Url}\fi

\bibitem[Achiam et~al.(2023)Achiam, Adler, Agarwal, Ahmad, Akkaya, Aleman, Almeida, Altenschmidt, Altman, Anadkat, et~al.]{achiam2023gpt4}
Josh Achiam, Steven Adler, Sandhini Agarwal, Lama Ahmad, Ilge Akkaya, Florencia~Leoni Aleman, Diogo Almeida, Janko Altenschmidt, Sam Altman, Shyamal Anadkat, et~al.
\newblock Gpt-4 technical report.
\newblock \emph{arXiv preprint arXiv:2303.08774}, 2023.

\bibitem[Alayrac et~al.(2022)Alayrac, Donahue, Luc, Miech, Barr, Hasson, Lenc, Mensch, Millican, Reynolds, et~al.]{alayrac2022flamingo}
Jean-Baptiste Alayrac, Jeff Donahue, Pauline Luc, Antoine Miech, Iain Barr, Yana Hasson, Karel Lenc, Arthur Mensch, Katherine Millican, Malcolm Reynolds, et~al.
\newblock Flamingo: a visual language model for few-shot learning.
\newblock In \emph{Advances in neural information processing systems}, pages 23716--23736, 2022.

\bibitem[Awadalla et~al.(2023)Awadalla, Gao, Gardner, Hessel, Hanafy, Zhu, Marathe, Bitton, Gadre, Sagawa, et~al.]{openflamingo}
Anas Awadalla, Irena Gao, Josh Gardner, Jack Hessel, Yusuf Hanafy, Wanrong Zhu, Kalyani Marathe, Yonatan Bitton, Samir Gadre, Shiori Sagawa, et~al.
\newblock Openflamingo: An open-source framework for training large autoregressive vision-language models.
\newblock \emph{arXiv preprint arXiv:2308.01390}, 2023.

\bibitem[Bai et~al.(2023)Bai, Bai, Yang, Wang, Tan, Wang, Lin, Zhou, and Zhou]{bai2023qwen}
Jinze Bai, Shuai Bai, Shusheng Yang, Shijie Wang, Sinan Tan, Peng Wang, Junyang Lin, Chang Zhou, and Jingren Zhou.
\newblock Qwen-vl: A versatile vision-language model for understanding, localization, text reading, and beyond.
\newblock 2023.

\bibitem[Changpinyo et~al.(2021)Changpinyo, Sharma, Ding, and Soricut]{cc12m}
Soravit Changpinyo, Piyush Sharma, Nan Ding, and Radu Soricut.
\newblock Conceptual 12m: Pushing web-scale image-text pre-training to recognize long-tail visual concepts.
\newblock In \emph{Proceedings of the IEEE/CVF conference on computer vision and pattern recognition}, pages 3558--3568, 2021.

\bibitem[Chen et~al.(2023{\natexlab{a}})Chen, Zhang, Zeng, Zhang, Zhu, and Zhao]{chen2023shikra}
Keqin Chen, Zhao Zhang, Weili Zeng, Richong Zhang, Feng Zhu, and Rui Zhao.
\newblock Shikra: Unleashing multimodal llm's referential dialogue magic.
\newblock \emph{arXiv preprint arXiv:2306.15195}, 2023{\natexlab{a}}.

\bibitem[Chen et~al.(2023{\natexlab{b}})Chen, Li, Dong, Zhang, He, Wang, Zhao, and Lin]{chen2023sharegpt4v}
Lin Chen, Jisong Li, Xiaoyi Dong, Pan Zhang, Conghui He, Jiaqi Wang, Feng Zhao, and Dahua Lin.
\newblock Sharegpt4v: Improving large multi-modal models with better captions.
\newblock \emph{arXiv preprint arXiv:2311.12793}, 2023{\natexlab{b}}.

\bibitem[Chen et~al.(2023{\natexlab{c}})Chen, Chen, Wang, Zhang, Yao, Feng, Han, Ding, Zeng, and Wang]{chen2023group}
Qiang Chen, Xiaokang Chen, Jian Wang, Shan Zhang, Kun Yao, Haocheng Feng, Junyu Han, Errui Ding, Gang Zeng, and Jingdong Wang.
\newblock Group detr: Fast detr training with group-wise one-to-many assignment.
\newblock In \emph{Proceedings of the IEEE/CVF International Conference on Computer Vision}, pages 6633--6642, 2023{\natexlab{c}}.

\bibitem[Chen et~al.(2015)Chen, Fang, Lin, Vedantam, Gupta, Doll{\'a}r, and Zitnick]{chen2015cococap}
Xinlei Chen, Hao Fang, Tsung-Yi Lin, Ramakrishna Vedantam, Saurabh Gupta, Piotr Doll{\'a}r, and C~Lawrence Zitnick.
\newblock Microsoft coco captions: Data collection and evaluation server.
\newblock \emph{arXiv preprint arXiv:1504.00325}, 2015.

\bibitem[Chen et~al.(2024)Chen, Wu, Wang, Su, Chen, Xing, Zhong, Zhang, Zhu, Lu, et~al.]{chen2024internvl}
Zhe Chen, Jiannan Wu, Wenhai Wang, Weijie Su, Guo Chen, Sen Xing, Muyan Zhong, Qinglong Zhang, Xizhou Zhu, Lewei Lu, et~al.
\newblock Internvl: Scaling up vision foundation models and aligning for generic visual-linguistic tasks.
\newblock In \emph{Proceedings of the IEEE/CVF Conference on Computer Vision and Pattern Recognition}, pages 24185--24198, 2024.

\bibitem[Chiang et~al.(2023)Chiang, Li, Lin, Sheng, Wu, Zhang, Zheng, Zhuang, Zhuang, Gonzalez, et~al.]{chiang2023vicuna}
Wei-Lin Chiang, Zhuohan Li, Zi Lin, Ying Sheng, Zhanghao Wu, Hao Zhang, Lianmin Zheng, Siyuan Zhuang, Yonghao Zhuang, Joseph~E Gonzalez, et~al.
\newblock Vicuna: An open-source chatbot impressing gpt-4 with 90\%* chatgpt quality.
\newblock \emph{See https://vicuna. lmsys. org (accessed 14 April 2023)}, 2\penalty0 (3):\penalty0 6, 2023.

\bibitem[Dai et~al.(2024)Dai, Li, Li, Tiong, Zhao, Wang, Li, Fung, and Hoi]{dai2024instructblip}
Wenliang Dai, Junnan Li, Dongxu Li, Anthony Meng~Huat Tiong, Junqi Zhao, Weisheng Wang, Boyang Li, Pascale~N Fung, and Steven Hoi.
\newblock Instructblip: Towards general-purpose vision-language models with instruction tuning.
\newblock \emph{Advances in Neural Information Processing Systems}, 36, 2024.

\bibitem[Du et~al.(2025)Du, Wang, Sun, Wang, Liao, Zhang, Ding, Wang, Wang, and Liu]{du2025lami}
Penghui Du, Yu Wang, Yifan Sun, Luting Wang, Yue Liao, Gang Zhang, Errui Ding, Yan Wang, Jingdong Wang, and Si Liu.
\newblock Lami-detr: Open-vocabulary detection with language model instruction.
\newblock In \emph{European Conference on Computer Vision}, pages 312--328, 2025.

\bibitem[Feng and Lapata(2012)]{feng2012automatic}
Yansong Feng and Mirella Lapata.
\newblock Automatic caption generation for news images.
\newblock \emph{IEEE Transactions on Pattern Analysis and Machine Intelligence}, 35\penalty0 (4):\penalty0 797--812, 2012.

\bibitem[Gao et~al.(2023)Gao, Han, Zhang, Lin, Geng, Zhou, Zhang, Lu, He, Yue, et~al.]{gao2023llamaadapter2}
Peng Gao, Jiaming Han, Renrui Zhang, Ziyi Lin, Shijie Geng, Aojun Zhou, Wei Zhang, Pan Lu, Conghui He, Xiangyu Yue, et~al.
\newblock Llama-adapter v2: Parameter-efficient visual instruction model.
\newblock \emph{arXiv preprint arXiv:2304.15010}, 2023.

\bibitem[Goyal et~al.(2017)Goyal, Khot, Summers-Stay, Batra, and Parikh]{goyal2017vqav2}
Yash Goyal, Tejas Khot, Douglas Summers-Stay, Dhruv Batra, and Devi Parikh.
\newblock Making the v in vqa matter: Elevating the role of image understanding in visual question answering.
\newblock In \emph{Proceedings of the IEEE conference on computer vision and pattern recognition}, pages 6904--6913, 2017.

\bibitem[Guo et~al.(2024)Guo, Jiao, Shen, Nie, and Kankanhalli]{guo2024unk}
Yangyang Guo, Fangkai Jiao, Zhiqi Shen, Liqiang Nie, and Mohan Kankanhalli.
\newblock Unk-vqa: A dataset and a probe into the abstention ability of multi-modal large models.
\newblock \emph{IEEE Transactions on Pattern Analysis and Machine Intelligence}, 46\penalty0 (12):\penalty0 10284--10296, 2024.

\bibitem[Hudson and Manning(2019)]{hudson2019gqa}
Drew~A Hudson and Christopher~D Manning.
\newblock Gqa: A new dataset for real-world visual reasoning and compositional question answering.
\newblock In \emph{Proceedings of the IEEE conference on computer vision and pattern recognition}, pages 6700--6709, 2019.

\bibitem[Ji et~al.(2022)Ji, Ma, Sun, Zhou, Wu, and Ji]{ji2022knowing}
Jiayi Ji, Yiwei Ma, Xiaoshuai Sun, Yiyi Zhou, Yongjian Wu, and Rongrong Ji.
\newblock Knowing what to learn: a metric-oriented focal mechanism for image captioning.
\newblock \emph{IEEE Transactions on Image Processing}, 31:\penalty0 4321--4335, 2022.

\bibitem[Joseph et~al.(2021)Joseph, Khan, Khan, and Balasubramanian]{joseph2021towards}
KJ Joseph, Salman Khan, Fahad~Shahbaz Khan, and Vineeth~N Balasubramanian.
\newblock Towards open world object detection.
\newblock In \emph{Proceedings of the IEEE conference on computer vision and pattern recognition}, pages 5830--5840, 2021.

\bibitem[Kembhavi et~al.(2016)Kembhavi, Salvato, Kolve, Seo, Hajishirzi, and Farhadi]{ai2d}
Aniruddha Kembhavi, Mike Salvato, Eric Kolve, Minjoon Seo, Hannaneh Hajishirzi, and Ali Farhadi.
\newblock A diagram is worth a dozen images.
\newblock In \emph{European Conference on Computer Vision}, pages 235--251, 2016.

\bibitem[Krishna et~al.(2017)Krishna, Zhu, Groth, Johnson, Hata, Kravitz, Chen, Kalantidis, Li, Shamma, et~al.]{krishna2017VG}
Ranjay Krishna, Yuke Zhu, Oliver Groth, Justin Johnson, Kenji Hata, Joshua Kravitz, Stephanie Chen, Yannis Kalantidis, Li-Jia Li, David~A Shamma, et~al.
\newblock Visual genome: Connecting language and vision using crowdsourced dense image annotations.
\newblock \emph{International journal of computer vision}, 123:\penalty0 32--73, 2017.

\bibitem[Lauren{\c{c}}on et~al.(2024)Lauren{\c{c}}on, Saulnier, Tronchon, Bekman, Singh, Lozhkov, Wang, Karamcheti, Rush, Kiela, et~al.]{laurenccon2024obelics}
Hugo Lauren{\c{c}}on, Lucile Saulnier, L{\'e}o Tronchon, Stas Bekman, Amanpreet Singh, Anton Lozhkov, Thomas Wang, Siddharth Karamcheti, Alexander Rush, Douwe Kiela, et~al.
\newblock Obelics: An open web-scale filtered dataset of interleaved image-text documents.
\newblock In \emph{Advances in Neural Information Processing Systems}, 2024.

\bibitem[Li et~al.(2023{\natexlab{a}})Li, Wang, Wang, Ge, Ge, and Shan]{li2023seed-bench}
Bohao Li, Rui Wang, Guangzhi Wang, Yuying Ge, Yixiao Ge, and Ying Shan.
\newblock Seed-bench: Benchmarking multimodal llms with generative comprehension.
\newblock \emph{arXiv preprint arXiv:2307.16125}, 2023{\natexlab{a}}.

\bibitem[Li et~al.(2023{\natexlab{b}})Li, Zhang, Chen, Wang, Yang, and Liu]{li2023otter}
Bo Li, Yuanhan Zhang, Liangyu Chen, Jinghao Wang, Jingkang Yang, and Ziwei Liu.
\newblock Otter: A multi-modal model with in-context instruction tuning.
\newblock \emph{arXiv preprint arXiv:2305.03726}, 2023{\natexlab{b}}.

\bibitem[Li et~al.(2024{\natexlab{a}})Li, Zhang, Zhang, Guo, Zhang, Zhang, Li, Liu, and Li]{coco-llava}
Bo Li, Hao Zhang, Kaichen Zhang, Dong Guo, Yuanhan Zhang, Renrui Zhang, Feng Li, Ziwei Liu, and Chunyuan Li.
\newblock Llava-next: What else influences visual instruction tuning beyond data?, 2024{\natexlab{a}}.

\bibitem[Li et~al.(2024{\natexlab{b}})Li, Zhang, Zhang, Guo, Zhang, Li, Zhang, Liu, and Li]{li2024llavanext-strong}
Bo Li, Kaichen Zhang, Hao Zhang, Dong Guo, Renrui Zhang, Feng Li, Yuanhan Zhang, Ziwei Liu, and Chunyuan Li.
\newblock Llava-next: Stronger llms supercharge multimodal capabilities in the wild, 2024{\natexlab{b}}.

\bibitem[Li et~al.(2022)Li, Li, Xiong, and Hoi]{li2022blip}
Junnan Li, Dongxu Li, Caiming Xiong, and Steven Hoi.
\newblock Blip: Bootstrapping language-image pre-training for unified vision-language understanding and generation.
\newblock In \emph{International conference on machine learning}, pages 12888--12900. PMLR, 2022.

\bibitem[Li et~al.(2024{\natexlab{c}})Li, Zhang, Diao, Wang, Wang, and Duan]{Densefusion}
Xiaotong Li, Fan Zhang, Haiwen Diao, Yueze Wang, Xinlong Wang, and Ling-Yu Duan.
\newblock Densefusion-1m: Merging vision experts for comprehensive multimodal perception.
\newblock \emph{arXiv preprint arXiv:2407.08303}, 2024{\natexlab{c}}.

\bibitem[Li et~al.(2023{\natexlab{c}})Li, Du, Zhou, Wang, Zhao, and Wen]{li2023pope}
Yifan Li, Yifan Du, Kun Zhou, Jinpeng Wang, Xin Zhao, and Ji-Rong Wen.
\newblock Evaluating object hallucination in large vision-language models.
\newblock In \emph{The 2023 Conference on Empirical Methods in Natural Language Processing}, 2023{\natexlab{c}}.

\bibitem[Li et~al.(2024{\natexlab{d}})Li, Yang, Liu, Ma, Zhang, Yang, Sun, Liu, and Bai]{li2024monkey}
Zhang Li, Biao Yang, Qiang Liu, Zhiyin Ma, Shuo Zhang, Jingxu Yang, Yabo Sun, Yuliang Liu, and Xiang Bai.
\newblock Monkey: Image resolution and text label are important things for large multi-modal models.
\newblock In \emph{Proceedings of the IEEE Conference on Computer Vision and Pattern Recognition}, pages 26763--26773, 2024{\natexlab{d}}.

\bibitem[Lin et~al.(2014{\natexlab{a}})Lin, Maire, Belongie, Hays, Perona, Ramanan, Doll{\'a}r, and Zitnick]{coco}
Tsung-Yi Lin, Michael Maire, Serge Belongie, James Hays, Pietro Perona, Deva Ramanan, Piotr Doll{\'a}r, and C~Lawrence Zitnick.
\newblock Microsoft coco: Common objects in context.
\newblock In \emph{Computer Vision--ECCV 2014: 13th European Conference, Zurich, Switzerland, September 6-12, 2014, Proceedings, Part V 13}, pages 740--755. Springer, 2014{\natexlab{a}}.

\bibitem[Lin et~al.(2014{\natexlab{b}})Lin, Maire, Belongie, Hays, Perona, Ramanan, Doll{\'a}r, and Zitnick]{lin2014coco}
Tsung-Yi Lin, Michael Maire, Serge Belongie, James Hays, Pietro Perona, Deva Ramanan, Piotr Doll{\'a}r, and C~Lawrence Zitnick.
\newblock Microsoft coco: Common objects in context.
\newblock In \emph{European Conference on Computer Vision}, pages 740--755. Springer, 2014{\natexlab{b}}.

\bibitem[Liu et~al.(2023{\natexlab{a}})Liu, Li, Li, and Lee]{liu2023llava1.5}
Haotian Liu, Chunyuan Li, Yuheng Li, and Yong~Jae Lee.
\newblock Improved baselines with visual instruction tuning.
\newblock \emph{arXiv preprint arXiv:2310.03744}, 2023{\natexlab{a}}.

\bibitem[Liu et~al.(2024{\natexlab{a}})Liu, Li, Li, Li, Zhang, Shen, and Lee]{liu2024llavanext_15}
Haotian Liu, Chunyuan Li, Yuheng Li, Bo Li, Yuanhan Zhang, Sheng Shen, and Yong~Jae Lee.
\newblock Llava-next: Improved reasoning, ocr, and world knowledge, 2024{\natexlab{a}}.

\bibitem[Liu et~al.(2024{\natexlab{b}})Liu, Li, Wu, and Lee]{llava}
Haotian Liu, Chunyuan Li, Qingyang Wu, and Yong~Jae Lee.
\newblock Visual instruction tuning.
\newblock In \emph{Advances in neural information processing systems}, 2024{\natexlab{b}}.

\bibitem[Liu et~al.(2023{\natexlab{b}})Liu, Duan, Zhang, Li, Zhang, Zhao, Yuan, Wang, He, Liu, et~al.]{liu2023mmbench}
Yuan Liu, Haodong Duan, Yuanhan Zhang, Bo Li, Songyang Zhang, Wangbo Zhao, Yike Yuan, Jiaqi Wang, Conghui He, Ziwei Liu, et~al.
\newblock Mmbench: Is your multi-modal model an all-around player?
\newblock \emph{arXiv preprint arXiv:2307.06281}, 2023{\natexlab{b}}.

\bibitem[Lu et~al.(2022)Lu, Mishra, Xia, Qiu, Chang, Zhu, Tafjord, Clark, and Kalyan]{scienceQA}
Pan Lu, Swaroop Mishra, Tanglin Xia, Liang Qiu, Kai-Wei Chang, Song-Chun Zhu, Oyvind Tafjord, Peter Clark, and Ashwin Kalyan.
\newblock Learn to explain: Multimodal reasoning via thought chains for science question answering.
\newblock In \emph{Advances in Neural Information Processing Systems}, pages 2507--2521, 2022.

\bibitem[Luo et~al.(2024)Luo, Zhou, Zhang, Zheng, Sun, and Ji]{luo2024feast}
Gen Luo, Yiyi Zhou, Yuxin Zhang, Xiawu Zheng, Xiaoshuai Sun, and Rongrong Ji.
\newblock Feast your eyes: Mixture-of-resolution adaptation for multimodal large language models.
\newblock \emph{arXiv preprint arXiv:2403.03003}, 2024.

\bibitem[Mathew et~al.(2021)Mathew, Karatzas, and Jawahar]{mathew2021docvqa}
Minesh Mathew, Dimosthenis Karatzas, and CV Jawahar.
\newblock Docvqa: A dataset for vqa on document images.
\newblock In \emph{Proceedings of the IEEE winter conference on applications of computer vision}, pages 2200--2209, 2021.

\bibitem[Meng et~al.(2021)Meng, Chen, Fan, Zeng, Li, Yuan, Sun, and Wang]{meng2021conditional}
Depu Meng, Xiaokang Chen, Zejia Fan, Gang Zeng, Houqiang Li, Yuhui Yuan, Lei Sun, and Jingdong Wang.
\newblock Conditional detr for fast training convergence.
\newblock In \emph{Proceedings of the IEEE/CVF international conference on computer vision}, pages 3651--3660, 2021.

\bibitem[Min et~al.(2023)Min, Wang, Liu, Luo, Kang, Wei, Wei, and Jiang]{prenet}
Weiqing Min, Zhiling Wang, Yuxin Liu, Mengjiang Luo, Liping Kang, Xiaoming Wei, Xiaolin Wei, and Shuqiang Jiang.
\newblock Large scale visual food recognition.
\newblock \emph{IEEE Transactions on Pattern Analysis and Machine Intelligence}, 45\penalty0 (8):\penalty0 9932--9949, 2023.

\bibitem[Onoe et~al.(2024)Onoe, Rane, Berger, Bitton, Cho, Garg, Ku, Parekh, Pont-Tuset, Tanzer, et~al.]{DOCCI}
Yasumasa Onoe, Sunayana Rane, Zachary Berger, Yonatan Bitton, Jaemin Cho, Roopal Garg, Alexander Ku, Zarana Parekh, Jordi Pont-Tuset, Garrett Tanzer, et~al.
\newblock Docci: Descriptions of connected and contrasting images.
\newblock \emph{arXiv preprint arXiv:2404.19753}, 2024.

\bibitem[Ordonez et~al.(2011)Ordonez, Kulkarni, and Berg]{ordonez2011sbucap}
Vicente Ordonez, Girish Kulkarni, and Tamara Berg.
\newblock Im2text: Describing images using 1 million captioned photographs.
\newblock In \emph{Advances in Neural Information Processing Systems}, 2011.

\bibitem[Peng et~al.(2023)Peng, Wang, Dong, Hao, Huang, Ma, and Wei]{peng2023kosmos}
Zhiliang Peng, Wenhui Wang, Li Dong, Yaru Hao, Shaohan Huang, Shuming Ma, and Furu Wei.
\newblock Kosmos-2: Grounding multimodal large language models to the world.
\newblock \emph{arXiv preprint arXiv:2306.14824}, 2023.

\bibitem[Pi et~al.(2024)Pi, Zhang, Zhang, Pan, Chen, and Zhang]{image_Tex}
Renjie Pi, Jianshu Zhang, Jipeng Zhang, Rui Pan, Zhekai Chen, and Tong Zhang.
\newblock Image textualization: An automatic framework for creating accurate and detailed image descriptions.
\newblock \emph{arXiv preprint arXiv:2406.07502}, 2024.

\bibitem[Plummer et~al.(2015)Plummer, Wang, Cervantes, Caicedo, Hockenmaier, and Lazebnik]{flickr30k}
Bryan~A Plummer, Liwei Wang, Chris~M Cervantes, Juan~C Caicedo, Julia Hockenmaier, and Svetlana Lazebnik.
\newblock Flickr30k entities: Collecting region-to-phrase correspondences for richer image-to-sentence models.
\newblock In \emph{Proceedings of the IEEE international conference on computer vision}, pages 2641--2649, 2015.

\bibitem[Radford et~al.(2021)Radford, Kim, Hallacy, Ramesh, Goh, Agarwal, Sastry, Askell, Mishkin, Clark, et~al.]{radford2021learning}
Alec Radford, Jong~Wook Kim, Chris Hallacy, Aditya Ramesh, Gabriel Goh, Sandhini Agarwal, Girish Sastry, Amanda Askell, Pamela Mishkin, Jack Clark, et~al.
\newblock Learning transferable visual models from natural language supervision.
\newblock In \emph{International conference on machine learning}, pages 8748--8763. PMLR, 2021.

\bibitem[Savchenko(2023)]{savchenko2023facial}
Andrey Savchenko.
\newblock Facial expression recognition with adaptive frame rate based on multiple testing correction.
\newblock In \emph{International Conference on Machine Learning}, pages 30119--30129, 2023.

\bibitem[Schuhmann et~al.(2022{\natexlab{a}})Schuhmann, Beaumont, Vencu, Gordon, Wightman, Cherti, Coombes, Katta, Mullis, Wortsman, et~al.]{laion5b}
Christoph Schuhmann, Romain Beaumont, Richard Vencu, Cade Gordon, Ross Wightman, Mehdi Cherti, Theo Coombes, Aarush Katta, Clayton Mullis, Mitchell Wortsman, et~al.
\newblock Laion-5b: An open large-scale dataset for training next generation image-text models.
\newblock \emph{Advances in Neural Information Processing Systems}, 35:\penalty0 25278--25294, 2022{\natexlab{a}}.

\bibitem[Schuhmann et~al.(2022{\natexlab{b}})Schuhmann, Beaumont, Vencu, Gordon, Wightman, Cherti, Coombes, Katta, Mullis, Wortsman, et~al.]{schuhmann2022laion}
Christoph Schuhmann, Romain Beaumont, Richard Vencu, Cade Gordon, Ross Wightman, Mehdi Cherti, Theo Coombes, Aarush Katta, Clayton Mullis, Mitchell Wortsman, et~al.
\newblock Laion-5b: An open large-scale dataset for training next generation image-text models.
\newblock In \emph{Advances in Neural Information Processing Systems}, pages 25278--25294, 2022{\natexlab{b}}.

\bibitem[Schwenk et~al.(2022)Schwenk, Khandelwal, Clark, Marino, and Mottaghi]{schwenk2022okvqa}
Dustin Schwenk, Apoorv Khandelwal, Christopher Clark, Kenneth Marino, and Roozbeh Mottaghi.
\newblock A-okvqa: A benchmark for visual question answering using world knowledge.
\newblock In \emph{European Conference on Computer Vision}, pages 146--162. Springer, 2022.

\bibitem[Sharma et~al.(2018)Sharma, Ding, Goodman, and Soricut]{cc3m}
Piyush Sharma, Nan Ding, Sebastian Goodman, and Radu Soricut.
\newblock Conceptual captions: A cleaned, hypernymed, image alt-text dataset for automatic image captioning.
\newblock In \emph{Proceedings of the 56th Annual Meeting of the Association for Computational Linguistics (Volume 1: Long Papers)}, pages 2556--2565, 2018.

\bibitem[Shi et~al.(2025)Shi, Wu, Mao, Wang, and Darrell]{shi2025llavas2}
Baifeng Shi, Ziyang Wu, Maolin Mao, Xin Wang, and Trevor Darrell.
\newblock When do we not need larger vision models?
\newblock In \emph{European Conference on Computer Vision}, pages 444--462. Springer, 2025.

\bibitem[Singh et~al.(2019)Singh, Natarajan, Shah, Jiang, Chen, Batra, Parikh, and Rohrbach]{textvqa}
Amanpreet Singh, Vivek Natarajan, Meet Shah, Yu Jiang, Xinlei Chen, Dhruv Batra, Devi Parikh, and Marcus Rohrbach.
\newblock Towards vqa models that can read.
\newblock In \emph{Proceedings of the IEEE conference on computer vision and pattern recognition}, pages 8317--8326, 2019.

\bibitem[Stefanini et~al.(2022)Stefanini, Cornia, Baraldi, Cascianelli, Fiameni, and Cucchiara]{stefanini2022show}
Matteo Stefanini, Marcella Cornia, Lorenzo Baraldi, Silvia Cascianelli, Giuseppe Fiameni, and Rita Cucchiara.
\newblock From show to tell: A survey on deep learning-based image captioning.
\newblock \emph{IEEE Transactions on Pattern Analysis and Machine Intelligence}, 45\penalty0 (1):\penalty0 539--559, 2022.

\bibitem[Stevens et~al.(2024)Stevens, Wu, Thompson, Campolongo, Song, Carlyn, Dong, Dahdul, Stewart, Berger-Wolf, et~al.]{stevens2024bioclip}
Samuel Stevens, Jiaman Wu, Matthew~J Thompson, Elizabeth~G Campolongo, Chan~Hee Song, David~Edward Carlyn, Li Dong, Wasila~M Dahdul, Charles Stewart, Tanya Berger-Wolf, et~al.
\newblock Bioclip: A vision foundation model for the tree of life.
\newblock In \emph{Proceedings of the IEEE Conference on Computer Vision and Pattern Recognition}, pages 19412--19424, 2024.

\bibitem[Sun et~al.(2023)Sun, Yu, Cui, Zhang, Zhang, Wang, Gao, Liu, Huang, and Wang]{sun2023emu}
Quan Sun, Qiying Yu, Yufeng Cui, Fan Zhang, Xiaosong Zhang, Yueze Wang, Hongcheng Gao, Jingjing Liu, Tiejun Huang, and Xinlong Wang.
\newblock Emu: Generative pretraining in multimodality.
\newblock In \emph{International Conference on Learning Representations}, 2023.

\bibitem[Sun et~al.(2024)Sun, Zhang, Chen, Zhang, Sang, Zhang, Wang, and Li]{sun2024improving}
Yanpeng Sun, Huaxin Zhang, Qiang Chen, Xinyu Zhang, Nong Sang, Gang Zhang, Jingdong Wang, and Zechao Li.
\newblock Improving multi-modal large language model through boosting vision capabilities.
\newblock \emph{arXiv preprint arXiv:2410.13733}, 2024.

\bibitem[Touvron et~al.(2023)Touvron, Martin, Stone, Albert, Almahairi, Babaei, Bashlykov, Batra, Bhargava, Bhosale, et~al.]{touvron2023llama2}
Hugo Touvron, Louis Martin, Kevin Stone, Peter Albert, Amjad Almahairi, Yasmine Babaei, Nikolay Bashlykov, Soumya Batra, Prajjwal Bhargava, Shruti Bhosale, et~al.
\newblock Llama 2: Open foundation and fine-tuned chat models.
\newblock \emph{arXiv preprint arXiv:2307.09288}, 2023.

\bibitem[Tu et~al.(2024)Tu, Li, Su, Zha, and Huang]{tu2024smart}
Yunbin Tu, Liang Li, Li Su, Zheng-Jun Zha, and Qingming Huang.
\newblock Smart: Syntax-calibrated multi-aspect relation transformer for change captioning.
\newblock \emph{IEEE Transactions on Pattern Analysis and Machine Intelligence}, 46\penalty0 (7):\penalty0 4926--4943, 2024.

\bibitem[Urbanek et~al.(2024)Urbanek, Bordes, Astolfi, Williamson, Sharma, and Romero-Soriano]{DCI}
Jack Urbanek, Florian Bordes, Pietro Astolfi, Mary Williamson, Vasu Sharma, and Adriana Romero-Soriano.
\newblock A picture is worth more than 77 text tokens: Evaluating clip-style models on dense captions.
\newblock In \emph{Proceedings of the IEEE Conference on Computer Vision and Pattern Recognition}, pages 26700--26709, 2024.

\bibitem[Wang et~al.(2024)Wang, Bai, Tan, Wang, Fan, Bai, Chen, Liu, Wang, Ge, et~al.]{wang2024qwen2-vl}
Peng Wang, Shuai Bai, Sinan Tan, Shijie Wang, Zhihao Fan, Jinze Bai, Keqin Chen, Xuejing Liu, Jialin Wang, Wenbin Ge, et~al.
\newblock Qwen2-vl: Enhancing vision-language model's perception of the world at any resolution.
\newblock \emph{arXiv preprint arXiv:2409.12191}, 2024.

\bibitem[Wang et~al.(2023)Wang, Lv, Yu, Hong, Qi, Wang, Ji, Yang, Zhao, Song, et~al.]{wang2023cogvlm}
Weihan Wang, Qingsong Lv, Wenmeng Yu, Wenyi Hong, Ji Qi, Yan Wang, Junhui Ji, Zhuoyi Yang, Lei Zhao, Xixuan Song, et~al.
\newblock Cogvlm: Visual expert for pretrained language models.
\newblock \emph{arXiv preprint arXiv:2311.03079}, 2023.

\bibitem[Yang et~al.(2024{\natexlab{a}})Yang, Kang, Huang, Xu, Feng, and Zhao]{yang2024depth}
Lihe Yang, Bingyi Kang, Zilong Huang, Xiaogang Xu, Jiashi Feng, and Hengshuang Zhao.
\newblock Depth anything: Unleashing the power of large-scale unlabeled data.
\newblock In \emph{Proceedings of the IEEE/CVF Conference on Computer Vision and Pattern Recognition}, pages 10371--10381, 2024{\natexlab{a}}.

\bibitem[Yang et~al.(2024{\natexlab{b}})Yang, Kang, Huang, Zhao, Xu, Feng, and Zhao]{depth_anything_v2}
Lihe Yang, Bingyi Kang, Zilong Huang, Zhen Zhao, Xiaogang Xu, Jiashi Feng, and Hengshuang Zhao.
\newblock Depth anything v2.
\newblock \emph{arXiv:2406.09414}, 2024{\natexlab{b}}.

\bibitem[Ye et~al.(2023)Ye, Xu, Ye, Yan, Liu, Qian, Zhang, Huang, and Zhou]{ye2023mplug2}
Qinghao Ye, Haiyang Xu, Jiabo Ye, Ming Yan, Haowei Liu, Qi Qian, Ji Zhang, Fei Huang, and Jingren Zhou.
\newblock mplug-owl2: Revolutionizing multi-modal large language model with modality collaboration.
\newblock \emph{arXiv preprint arXiv:2311.04257}, 2023.

\bibitem[Yu et~al.(2019)Yu, Li, Yu, and Huang]{yu2019multimodal}
Jun Yu, Jing Li, Zhou Yu, and Qingming Huang.
\newblock Multimodal transformer with multi-view visual representation for image captioning.
\newblock \emph{IEEE Transactions on Circuits and Systems for Video Technology}, 30\penalty0 (12):\penalty0 4467--4480, 2019.

\bibitem[Yu et~al.(2023)Yu, Yang, Li, Wang, Lin, Liu, Wang, and Wang]{yu2023mmvet}
Weihao Yu, Zhengyuan Yang, Linjie Li, Jianfeng Wang, Kevin Lin, Zicheng Liu, Xinchao Wang, and Lijuan Wang.
\newblock Mm-vet: Evaluating large multimodal models for integrated capabilities.
\newblock \emph{arXiv preprint arXiv:2308.02490}, 2023.

\bibitem[Yuan et~al.(2023)Yuan, Zhang, Wang, Albanie, Pan, Feng, Jiang, Ni, Zhang, and Zhao]{Yuan2023RLIPv2}
Hangjie Yuan, Shiwei Zhang, Xiang Wang, Samuel Albanie, Yining Pan, Tao Feng, Jianwen Jiang, Dong Ni, Yingya Zhang, and Deli Zhao.
\newblock Rlipv2: Fast scaling of relational language-image pre-training.
\newblock In \emph{Proceedings of the IEEE International Conference on Computer Vision}, 2023.

\bibitem[Yue et~al.(2024)Yue, Ni, Zhang, Zheng, Liu, Zhang, Stevens, Jiang, Ren, Sun, et~al.]{yue2024mmmu}
Xiang Yue, Yuansheng Ni, Kai Zhang, Tianyu Zheng, Ruoqi Liu, Ge Zhang, Samuel Stevens, Dongfu Jiang, Weiming Ren, Yuxuan Sun, et~al.
\newblock Mmmu: A massive multi-discipline multimodal understanding and reasoning benchmark for expert agi.
\newblock In \emph{Proceedings of the IEEE Conference on Computer Vision and Pattern Recognition}, pages 9556--9567, 2024.

\bibitem[Zha et~al.(2019)Zha, Liu, Zhang, Zhang, and Wu]{zha2019context}
Zheng-Jun Zha, Daqing Liu, Hanwang Zhang, Yongdong Zhang, and Feng Wu.
\newblock Context-aware visual policy network for fine-grained image captioning.
\newblock \emph{IEEE Transactions on Pattern Analysis and Machine Intelligence}, 44\penalty0 (2):\penalty0 710--722, 2019.

\bibitem[Zhang et~al.(2021)Zhang, Li, Zhai, and Liu]{mmalnet}
Fan Zhang, Meng Li, Guisheng Zhai, and Yizhao Liu.
\newblock Multi-branch and multi-scale attention learning for fine-grained visual categorization.
\newblock In \emph{International Conference on MultiMedia Modeling}, pages 136--147, 2021.

\bibitem[Zhang et~al.(2024)Zhang, Huang, Jin, and Lu]{zhang2024vision}
Jingyi Zhang, Jiaxing Huang, Sheng Jin, and Shijian Lu.
\newblock Vision-language models for vision tasks: A survey.
\newblock \emph{IEEE Transactions on Pattern Analysis and Machine Intelligence}, 2024.

\bibitem[Zhu et~al.(2023{\natexlab{a}})Zhu, Chen, Shen, Li, and Elhoseiny]{zhu2023minigpt4}
Deyao Zhu, Jun Chen, Xiaoqian Shen, Xiang Li, and Mohamed Elhoseiny.
\newblock Minigpt-4: Enhancing vision-language understanding with advanced large language models.
\newblock \emph{arXiv preprint arXiv:2304.10592}, 2023{\natexlab{a}}.

\bibitem[Zhu et~al.(2023{\natexlab{b}})Zhu, He, and Wu]{zhu2023quantized}
Ke Zhu, Yin-Yin He, and Jianxin Wu.
\newblock Quantized feature distillation for network quantization.
\newblock In \emph{Proceedings of the AAAI Conference on Artificial Intelligence}, pages 11452--11460, 2023{\natexlab{b}}.

\bibitem[Zhu et~al.(2024)Zhu, Zhao, Ge, and Zhang]{zhu2024self}
Ke Zhu, Liang Zhao, Zheng Ge, and Xiangyu Zhang.
\newblock Self-supervised visual preference alignment.
\newblock In \emph{Proceedings of ACM International Conference on Multimedia}, pages 291--300, 2024.

\end{thebibliography}
}
\clearpage

\appendix

\section{More Details of EDC pipeline}

\subsection{The Specific Visual Expert Model}
In EDC, we employ different visual expert models to obtain object-level and relation-level attribute labels.  The main paper provides an overview of the process for obtaining different attributes and briefly introduces the types of Visual Specialists used. In Table~\ref{tab:detail_expert}, we will further elaborate on the off-the-shelf visual specialists employed in the EDC pipeline. Based on our experience, existing open-source OCR models exhibit suboptimal performance in accurately recognizing text in images. To enhance OCR accuracy, we leverage Baidu’s API~\footnote{https://ai.baidu.com/tech/ocr/general}, which offers superior text recognition capabilities. Additionally, for attributes such as logo, celebrity, and landmark recognition, the limited availability of labeled data poses a challenge. To mitigate this, we also utilize Baidu’s API~\footnote{https://ai.baidu.com/tech/imagerecognition} to better capture these attributes, ensuring more precise and comprehensive annotations. Meanwhile, as a simple and versatile framework, EDC allows users to flexibly select the most suitable visual specialist models based on their available resources.

\subsection{The Detail of Relation Attributes}
Relation Attributes play a crucial role in image captioning. Among them, HOI captures human-object interactions, while 2D Absolute Location describes an object's position within the image. Since such information is relatively sparse, we retain all relations in the captions. In contrast, 2D/3D Relative Location encodes spatial relationships between objects, which can become abundant and redundant as the number of objects increases. To address this, we retain only one dominant relationship per image for 2D/3D Relative Location, prioritizing the one with the greatest relative distance to ensure accuracy.

\subsection{The Prompt Templates of EDC}
To effectively transform extracted visual attributes into natural language, EDC designs two structured prompt templates that guide the LLM to convert discrete visual signals into coherent and readable descriptions. The first template focuses on object-level attributes, integrating multiple properties of each individual object. The second focuses on relation-level attributes, summarizing the relationships among different objects. The entire process is designed to mimic human perception: \textit{first describing what each object is, and then explaining how these objects relate to each other within the scene.}

\begin{figure*}[ht]
    \centering
    \includegraphics[width=1.\linewidth]{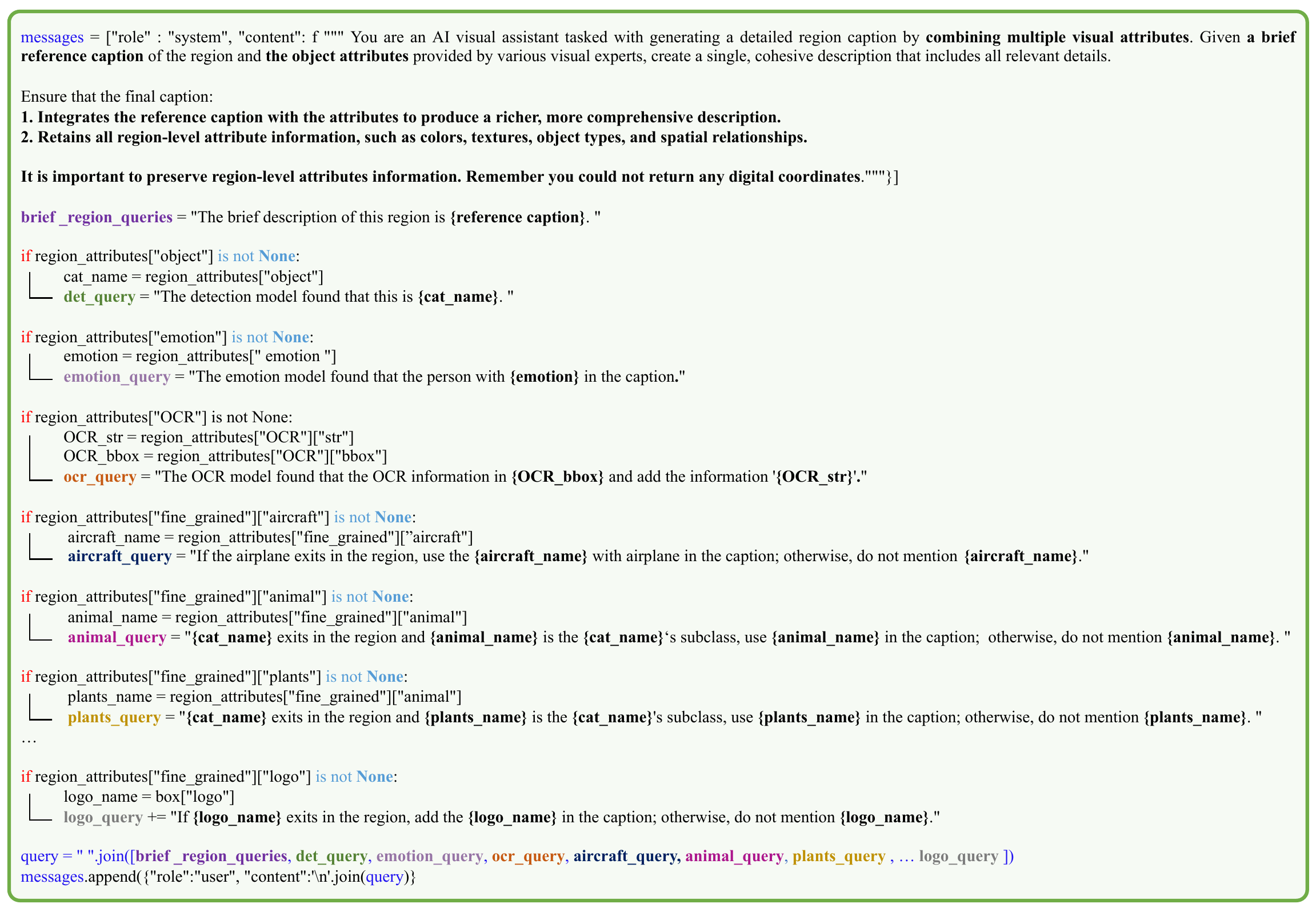}

    \caption{The prompt for using LLM to generate an region caption by considering object attributes and reference captions.}

    \label{fig:llm_prompt_object}
\end{figure*}

For the object-level prompt template (as shown in Figure~\ref{fig:llm_prompt_object}), the LLM uses structured attributes extracted by visual specialists—such as category, color, texture, emotion, fine-grained classification (\textit{e.g.}, replacing “dog” with “golden retriever”), OCR content, and depth—to check whether the reference caption already includes these details. If a specific attribute is missing, it will be added; if it conflicts with the extracted value, the information from the visual specialist takes priority. This process allows the model to merge all attributes into a richer region-level description without redundancy or hallucinated content. For example, the prompt explicitly defines conditional rules such as \textit{``If a fine-grained animal category is available, replace the coarse label with the specific species; otherwise, keep the original name.''} Likewise, it guides the LLM to naturally embed OCR text, detected emotions, and visual context into the caption, producing region-level descriptions that are both accurate and fluent.

\begin{figure*}[ht]
    \centering
    \includegraphics[width=1.\linewidth]{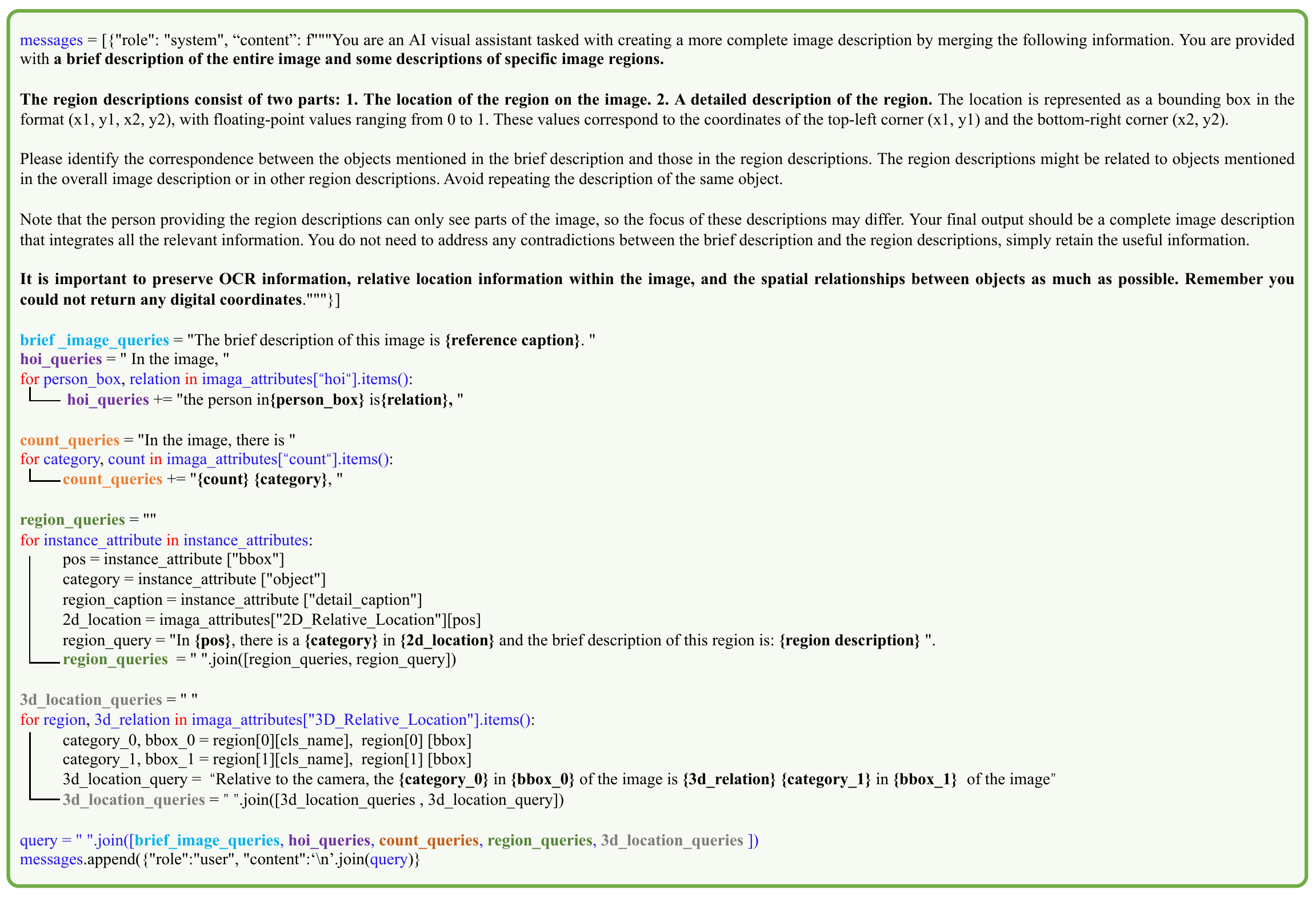}
    \caption{The prompt for LLM
    to generate an image caption
    by considering relation attributes, region location information and captions.}
    \label{fig:llm_prompt_relation}
\end{figure*}

For the relation-level prompt template (as shown in Figure~\ref{fig:llm_prompt_relation}), the LLM incorporates the relationships among different objects into the overall image caption. Among these, the P2O relations are directly obtained from the HOI model, while other types of relations are derived from detection and depth models. Specifically, the detection model provides the count of objects and, based on their bounding boxes, determines both 2D absolute locations and 2D relative locations to describe spatial relationships on the image plane. The depth model, on the other hand, extracts a depth value for each object from its bounding box and the depth map. By comparing the depth values between objects, the model can infer their 3D spatial relationships, such as whether object A is in front of or behind object B. Moreover, incorporating the object-level captions further enriches the image caption with detailed visual information. This structured prompting process ensures that the model maintains spatial grounding while merging scattered regional details into one coherent and comprehensive image-level description.

In practice, both templates are implemented as single-turn structured prompts rather than multi-turn reasoning. Each prompt directly guides the LLM to generate captions based on the provided visual attributes and contextual cues, without requiring iterative dialogue. The object-level template focuses on enriching region descriptions with fine-grained visual details, while the relation-level template integrates spatial and interaction information to form a complete image-level caption. Together, these two prompts enable EDC to seamlessly connect perception-level attributes with natural language expression, producing captions that are more detailed, precise, and contextually grounded than those from conventional LMMs.

\begin{table*}[!t]
\centering
\setlength\tabcolsep{4pt}
\renewcommand\arraystretch{1.3}
\setlength{\tabcolsep}{5.4mm}{
\vspace{-.8em}
\caption{The Specific Visual Expert Model of EDC.}

\label{tab:detail_expert}
\resizebox{\textwidth}{!}{%
\begin{tabular}{ccccccc}
\hline
\multicolumn{2}{c|}{\textbf{Detection Model}}                    & \multicolumn{2}{c|}{\multirow{2}{*}{\textbf{Depth Model}}} & \multicolumn{1}{c|}{\multirow{2}{*}{\textbf{OCR Model}}} & \multicolumn{1}{c|}{\multirow{2}{*}{\textbf{HoI Model}}} & \multirow{2}{*}{\textbf{Emotion Model}} \\ \cline{1-2}
\multicolumn{1}{c|}{In-domain}  & \multicolumn{1}{c|}{Open world} & \multicolumn{2}{c|}{}                                      & \multicolumn{1}{c|}{}                                    & \multicolumn{1}{c|}{}                                    &                                         \\ \hline
\multicolumn{1}{c|}{Group Detr~\cite{chen2023group}} & \multicolumn{1}{c|}{LaMI-DETR~\cite{du2025lami}}      & \multicolumn{2}{c|}{Depth Anything V2~\cite{depth_anything_v2}}                     & \multicolumn{1}{c|}{API$^1$}                                 & \multicolumn{1}{c|}{RLIPv2~\cite{Yuan2023RLIPv2}}                              & ~\cite{savchenko2023facial}                     \\ \hline
\multicolumn{7}{c}{\textbf{Fine-Grained Model}}                                                                                                                                                                                                                                              \\ \hline
\multicolumn{1}{c|}{Animal}     & \multicolumn{1}{c|}{Plant}     & \multicolumn{1}{c|}{Aircrafts} & \multicolumn{1}{c|}{Logo} & \multicolumn{1}{c|}{Landmark}                            & \multicolumn{1}{c|}{Food}                                & Celebrity                               \\ \hline
\multicolumn{1}{c|}{BioClip~\cite{stevens2024bioclip}}    & \multicolumn{1}{c|}{BioClip~\cite{stevens2024bioclip}}   & \multicolumn{1}{c|}{MMALNet~\cite{mmalnet}}   & \multicolumn{1}{c|}{API$^2$}  & \multicolumn{1}{c|}{API$^2$}                                 & \multicolumn{1}{c|}{PreNet~\cite{prenet}}                              & API$^2$                                     \\ \hline
\end{tabular}
}}

\end{table*}

\section{Analysis of EDC datasets}

The EDC dataset consists of two parts: EDC-1M, comprising 1 million diverse image-text pairs sampled from the Laion dataset~\cite{schuhmann2022laion}, and EDC-118K, comprising 118,000 real image-text pairs from the COCO dataset~\cite{coco}. Next, we will analyze the captions in EDC-118K and compare them with captions annotated by humans~\cite{chen2015cococap} and generic LMM models~\cite{chen2024internvl,li2024llavanext-strong}.

\subsection{The Caption Length}

In general, the longer caption could convey more detailed visual content. We compared the caption length of EDC-118k with human annotations as well as captions generated by advanced MLLM models, InternVL2-26B~\cite{chen2024internvl} and LLaVA-Next-34B~\cite{li2024llavanext-strong}. The results are summarized in Table~\ref{tab:data_pro}. It was observed that human-generated captions were the shortest, as they typically focus on only the most salient objects. The InternVL2-26B could generate more longer captions, with approximately 106 tokens. The captions generated by LLaVA-Next-34B were the longest, averaging around 228 tokens, while EDC-118k produced captions with an average of 218 tokens, approximately 10 tokens fewer than those of LLaVA-Next-34B.

\subsection{The Lexical Composition}

\begin{figure}[!ht]
    \centering
    \includegraphics[width=\linewidth]{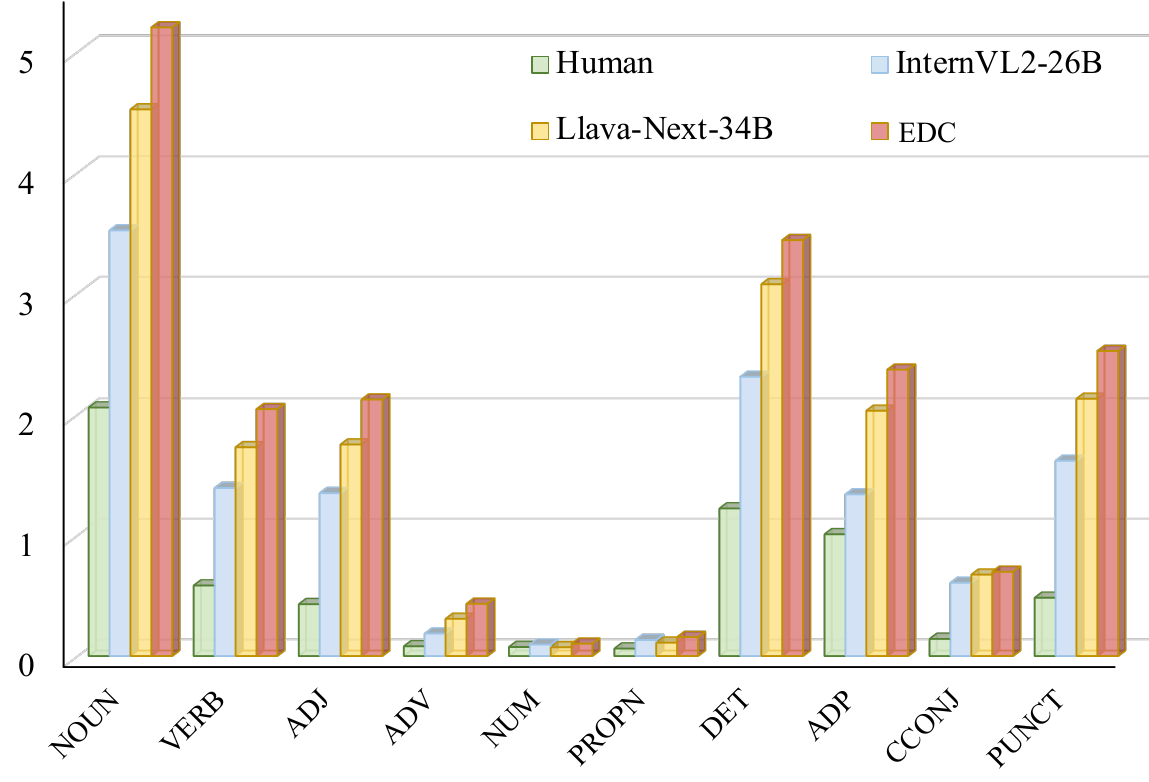}
    \caption{Comparison of lexical composition of the captions generated by different methods. The y-axis represents the average frequency of each class of lexical per sentence in the datasets.}
    \label{fig:lexical_com}
\end{figure}

We conducted a detailed analysis of the lexical composition of the captions to evaluate how effectively each model described the visual content. This analysis examined the variety, frequency, and distribution of different word categories, including nouns, verbs, adjectives, adverbs, numerals, and more. As shown in Fig.~\ref{fig:lexical_com}, EDC-118k contained the highest average number of lexical elements per sentence, demonstrating a more diverse and complex linguistic structure compared to other captions generated by other datasets. This richer composition indicates that EDC-118k was better equipped to deliver nuanced and detailed descriptions of visual scenes, using a wider range of grammatical constructs to convey more comprehensive information.

\begin{table}[]\footnotesize
    \caption{Comparison of the different caption datasets. The \textit{"ATL"} abbreviates the \textit{"Average Token Length"}. The token length is counted by the tokenlizer of Vicuna-v1.5}
 \label{tab:data_pro}
 \centering
 \renewcommand\arraystretch{1.4}
 \setlength{\tabcolsep}{2.5mm}{
 \resizebox{\linewidth}{!}{
\begin{tabular}{l|c|c|c}
\hline
 \textbf{Cpationed by} & \textbf{Image Source} & \textbf{Samples} & \textbf{ATL of Caption} \\ \hline
 Human & \multirow{4}{*}{COCO} & \multirow{4}{*}{118k} & 14.67 \\ \cline{1-1} \cline{4-4} 
  InternVL2-26B &  &  & 105.80 \\ \cline{1-1} \cline{4-4} 
  LLaVA-NeXT-34B &  &  & 227.68 \\ \cline{1-1} \cline{4-4} 
  EDC &  &  & 217.71 \\ \hline
\end{tabular}}}
\end{table}

\subsection{The Word Clouds}

\begin{figure*}[!ht]
    \centering
    \includegraphics[width=\linewidth]{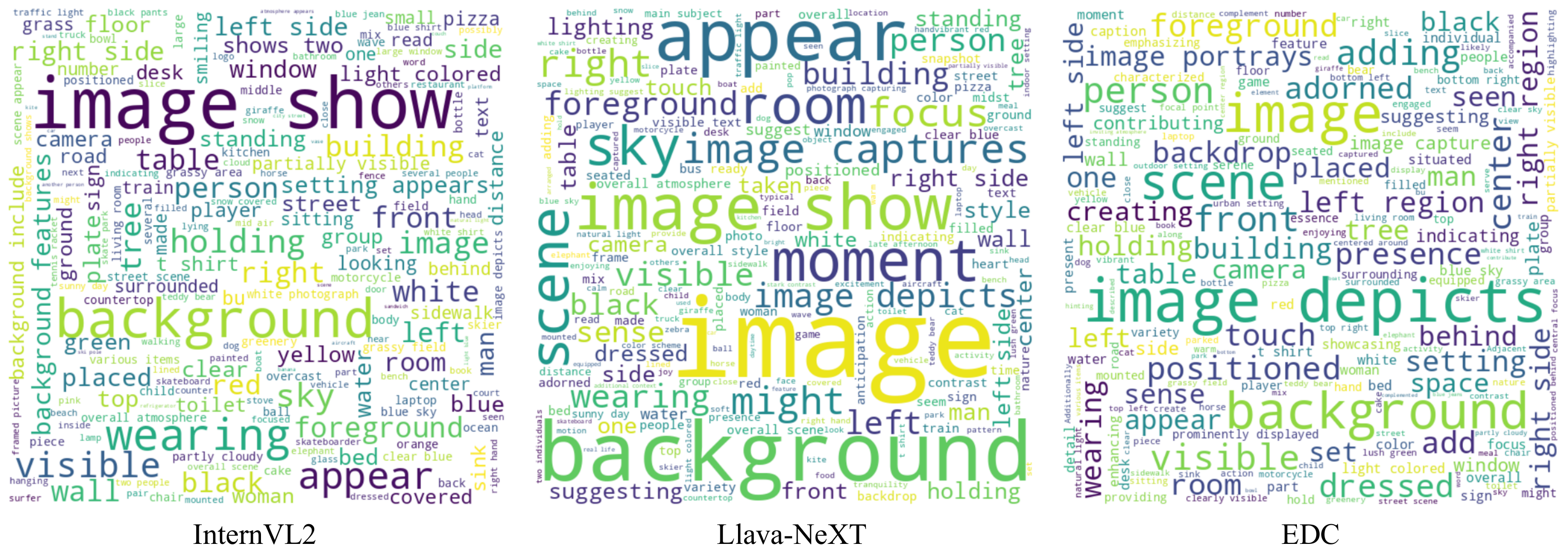}
    \caption{Word Cloud of captions generated by InternVL2, LLaVA-Next and EDC.}
    \label{fig:cloud}
\end{figure*}

In Fig.~\ref{fig:cloud}, we present word clouds for the captions generated by InternVL2-26B, LLaVA-Next-34B, and our EDC pipeline. These visualizations highlight the most frequently used words across the different captioning methods, providing an intuitive comparison of the lexical patterns and focus areas of each caption. By examining the word clouds, we observed that the captions generated by our EDC exhibited a notably diverse vocabulary. In particular, there was a significantly higher frequency of words describing the relative spatial relationships of objects, both in 2D and 3D space like \textit{`left side', `right region', `front'}, and \textit{`behind'}, compared to captions produced by other methods. This indicates that EDC not only captured a wider range of visual details but also excelled in conveying the spatial context of objects within the scenes, offering more comprehensive descriptions of the visual content.

\section{Training Details}

We elaborate the training details and hyper-parameters used in our experiments for evaluating the effectiveness of EDC-1M generated by our EDC pipeline. The whole training step consists of three stages, as shown in Table. \ref{tab: Training details}. During the pre-alignment stage, we exclusively train the projector, resulting in a more stable and consistent vision-language connection. In the pre-training phase, similar with ShareGPT4V~\cite{chen2023sharegpt4v}, we unfreeze the Vision Encoder (VE) for the last 12 layers, the Language Model (LM), and the projector. Regarding the instruction tuning stage, we use the open-source LLaVA-mix-665K~\cite{liu2023llava1.5} and LLaVA-NeXT-data to fine-tune both the projector and language model of the LLaVA-v1.5~\cite{liu2023llava1.5} and LLaVA-NeXT~\cite{liu2024llavanext_15} models, respectively.

\begin{table*}[!ht]
\centering
\caption{Training details and hyper-parameters used in our experiments. `VE' means the vision encoder of CLIP for the last 12 layers, and `LM' refers to the language model.}
\label{tab: Training details}
\renewcommand{\arraystretch}{1.2}
\setlength{\tabcolsep}{6.5mm}{
\resizebox{1.\linewidth}{!}{
    \begin{tabular}{lccc}
    \toprule
    \textbf{Hyper-parameter} & \textbf{Pre-aligning} & \textbf{Pre-training} & \textbf{Instruction Tuning}  \\
    \midrule
    Batch Size  & 256 & 256 & 128 \\
    Learning Rate   & 2e-5 & 2e-5 & 2e-5  \\
    LR Schedule  & \multicolumn{3}{c}{cosine decay}  \\
    LR Warmup Ratio  & 0.01 & 0.01 & 0.01  \\
    Weight Decay  & 0 & 0 & 0  \\
    \multirow{2}*{Trainable Module}  & \multirow{2}*{Projector} & \textbf{LLaVA-v1.5}: Projector, VE, LM & \textbf{LLaVA-v1.5}: Projector, LM \\
    &&\textbf{LLaVA-NeXT}: Full Model & \textbf{LLaVA-NeXT}: Full Model \\
    Epoch  & 1 & 1 & 1    \\
    Optimizer  & \multicolumn{3}{c}{AdamW} \\
    DeepSpeed stage  & 3 & 3 & 3  \\
    \multirow{2}*{Dataset} & \multirow{2}*{EDC-1M} & \multirow{2}*{EDC-1M} &\textbf{ LLaVA-v1.5}: LLaVA-mix-665K   \\
    ~ & & &  \textbf{LLaVA-NeXT}: LLaVA-NeXT-data \\
    \bottomrule
    \end{tabular}}}
\end{table*}

\begin{table*}[h]
\centering
\setlength\tabcolsep{4pt}
\renewcommand\arraystretch{1.4}
\setlength{\tabcolsep}{3.4mm}{
\caption{CircularEval multi-choice accuracy results on MMBench~\cite{liu2023mmbench} dev set.
We adopt the following abbreviations: LR for Logical
Reasoning; AR for Attribute Reasoning; RR for Relation Reasoning; FP-C for Fine-grained Perception (Cross Instance); FP-S for Finegrained
Perception (Single Instance); CP for Coarse Perception.}
\label{tab:mmbench}
\resizebox{\textwidth}{!}{%
\begin{tabular}{l|ccccccc|ccccccc}
\hline
 \multirow{2}{*}{\textbf{Annotation Method}} & \multicolumn{7}{c|}{\textbf{MMBench-CN}} & \multicolumn{7}{c}{\textbf{MMBench}} \\ \cline{2-15}
 & Overall & LR & AR & RR & FP-S & FP-C & CP & Overall & LR & AR & RR & FP-S & FP-C & CP \\ \hline
InternVL2-26B~\cite{chen2024internvl} & 56.9 & 28.8 & 58.8 & \textbf{59.1} & 54.9 & 44.1 & 74.0 & 64.8 & 35.6 & 68.3 & \textbf{57.4} & \textbf{70.3} & 52.4 & 77.4 \\ 
LLaVA-NeXT-34B~\cite{li2024llavanext-strong} & 56.5 & 28.8 & 60.3 & 56.5 & 54.3 & 42.0 & 74.3 & 64.9 & 31.4 & 68.3 & 57.4 & 69.3 & \textbf{54.5} & \textbf{79.7} \\ 
EDC & \textbf{58.2} & \textbf{29.7} & \textbf{62.3} & 57.4 & \textbf{56.7} & \textbf{45.5} & \textbf{74.7} & \textbf{65.8} & \textbf{37.3} & \textbf{71.4} & 57.4 & 70.0 & 53.8 & 78.4 \\ \arrayrulecolor{lightgray} \hline \arrayrulecolor{black} 
InternVL2-26B~\cite{chen2024internvl} & 58.8 & 31.4 & 60.3 & 51.3 & 56.0 & 51.0 & 78.4 & 66.7 & 36.4 & 71.4 & 59.1 & 66.9 & \textbf{62.9} & 80.1 \\ 
LLaVA-NeXT-34B~\cite{li2024llavanext-strong}  & 59.8 & 31.3 & 60.8 & 54.8 & 56.7 & 49.7 & \textbf{80.7} & 67.2 & 38.1 & \textbf{69.8} & 57.4 & \textbf{69.3} & 60.1 & 82.1 \\ 
EDC & \textbf{60.1} & \textbf{31.4} & \textbf{62.3} & \textbf{55.7} & \textbf{57.0} & \textbf{51.0} & 79.1 & \textbf{68.5} & \textbf{39.0} & \textbf{72.4} & \textbf{67.8} & 67.9 & 59.4 & \textbf{82.8} \\ \hline
\end{tabular}
}}
\end{table*}

\section{Detalied Evaluation Results}
To highlight the significant improvements EDC-generated image descriptions bring to model performance, we present a detailed evaluation of the MMbench results in Table~\ref{tab:mmbench}. These results demonstrate how high-quality image descriptions enhance the model's capabilities, particularly in logical reasoning, attribute reasoning, and relational reasoning. This improvement is driven by EDC’s ability to accurately capture object relationships and detailed attributes within images, enabling more effective reasoning and a deeper understanding of object interactions and characteristics.

Moreover, EDC-generated descriptions achieve competitive performance in both fine-grained and coarse perception tasks, showcasing the effectiveness of integrating various visual experts to emulate manual annotation. This approach enriches visual information, resulting in higher accuracy and robustness in tasks like visual question answering and image comprehension.

\section{Visualizations on EDC}
\begin{figure*}[!ht]
    \centering
    \vspace{-1.8em}
    \includegraphics[width=.88\linewidth]{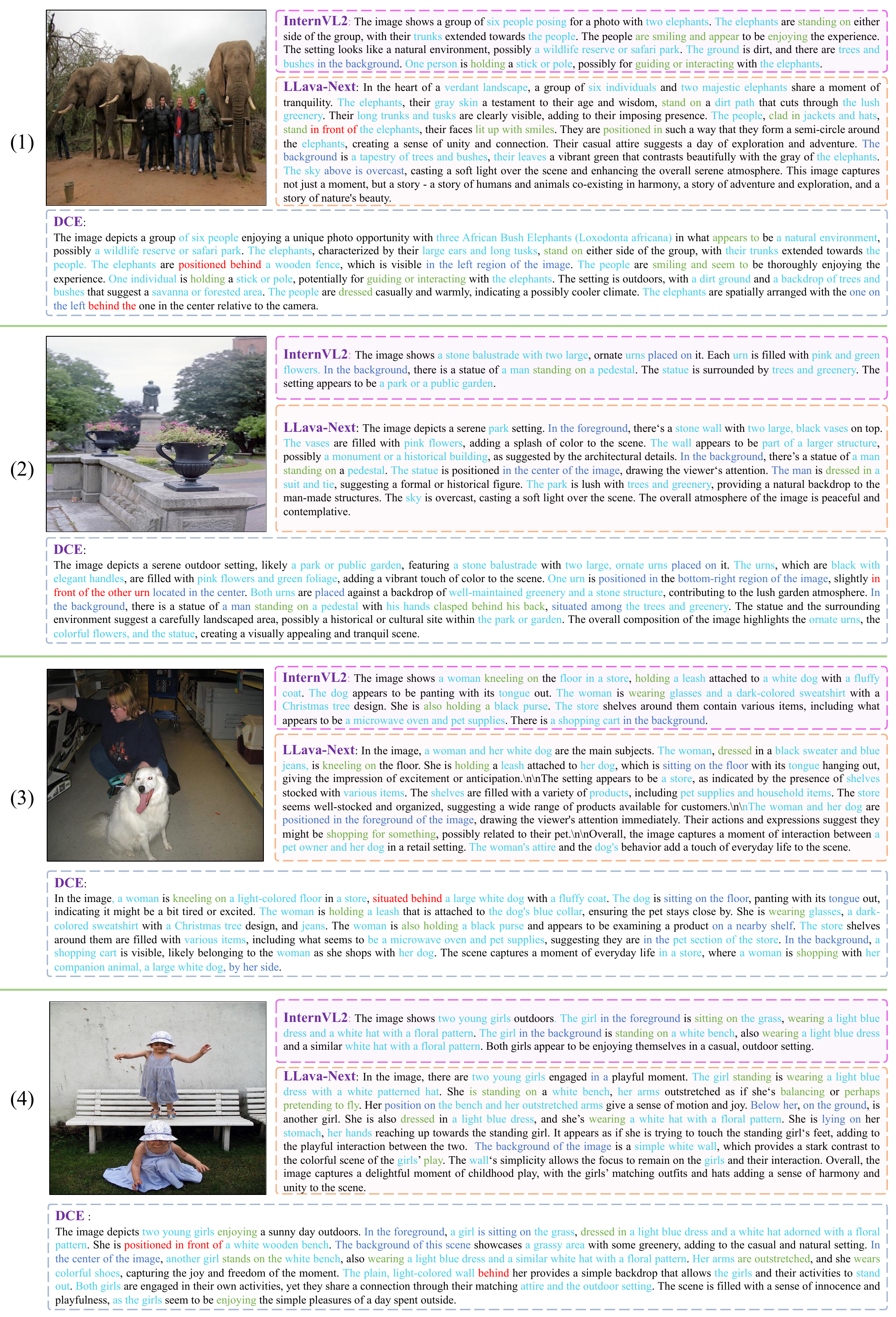}
    \caption{A comparison of image captions generated by InternVL2-26B, LLaVA-Next-34B, and EDC. We highlight different types of information, including \textcolor{object_skyblue}{Object Attributes}, \textcolor{ocr_coral}{OCR}, \textcolor{hoi_green}{HOI}, \textcolor{blue}{2D spatial relations} and \textcolor{red}{3D spatial relations}.}
    \label{fig:vis_appendix}
\end{figure*}

To visually demonstrate the quality of captions annotated by EDC, we compared them with captions generated by generic LMMs, such as InternVL2-26B and LLaVA-NeXT-34B. The visualization highlights the differences in caption quality, providing a clear comparison of EDC's detailed and accurate descriptions against those produced by the LMMs. As shown in Figure~\ref{fig:vis_appendix}, EDC captions consistently capture more nuanced object attributes, relationships, and contextual details, showcasing its superior annotation capabilities. This comparison underscores EDC’s effectiveness in generating high-quality captions that enhance downstream visual-language tasks.

\end{document}